\documentclass{article}
\usepackage{ijcai24}

\usepackage{times}
\usepackage{soul}
\usepackage{url}
\usepackage[hidelinks]{hyperref}
\usepackage[utf8]{inputenc}
\usepackage[small]{caption}
\usepackage{graphicx}
\usepackage{amsmath}
\usepackage{amsthm}
\usepackage{booktabs}
\usepackage{algorithm}
\usepackage{algorithmic}
\usepackage[switch]{lineno}

\usepackage{makecell}
\usepackage{amssymb,amsfonts,bbding}
\usepackage{textcomp}
\usepackage{bm}
\usepackage{multirow}
\usepackage{threeparttable}
\usepackage{color}
\usepackage{colortbl}
\usepackage{xcolor}
\usepackage{stfloats}
\usepackage{pifont}
\usepackage{utfsym}

\usepackage{tabularx} %导入tabularx包
\newcolumntype{Y}{>{\centering\arraybackslash}X}%将单元格类型改为自动换行模式

\title{SkyEyeGPT: Unifying Remote Sensing Vision-Language Tasks via Instruction Tuning with Large Language Model}

\author{
Yang Zhan$^1$ \and
Zhitong Xiong$^{2}$\and
Yuan Yuan$^1$ 
    \affiliations
    $^1$iOPEN, Northwestern Polytechnical University, Xi'an, China\\
$^2$Technical University of Munich (TUM), Munich, Germany\\
    \emails
   \{zhanyangnwpu, xiongzhitong, y.yuan1.ieee\}@gmail.com
}

\begin{document}

\maketitle

\begin{abstract}
Large language models (LLMs) have recently been extended to the vision-language realm, obtaining impressive general multi-modal capabilities. However, the exploration of multi-modal large language models (MLLMs) for remote sensing (RS) data is still in its infancy, and the performance is not satisfactory.
In this work, we introduce \textbf{SkyEyeGPT}, a unified multi-modal large language model specifically designed for RS vision-language understanding. To this end, we meticulously curate an RS multi-modal instruction tuning dataset, including single-task and multi-task conversation instructions. After manual verification, we obtain a high-quality RS instruction-following dataset with \textbf{968k} samples.
Our research demonstrates that with a simple yet effective design, SkyEyeGPT works surprisingly well on considerably different tasks without the need for extra encoding modules. Specifically, after projecting RS visual features to the language domain via an alignment layer, they are fed jointly with task-specific instructions into an LLM-based RS decoder to predict answers for RS open-ended tasks.
In addition, we design a two-stage tuning method to enhance instruction-following and multi-turn dialogue ability at different granularities. 
Experiments on \textbf{8} datasets for RS vision-language tasks demonstrate SkyEyeGPT's superiority in image-level and region-level tasks, such as captioning and visual grounding. In particular, SkyEyeGPT exhibits encouraging results compared to GPT-4V in some qualitative tests. The online demo, code, and dataset will be released.

\end{abstract}

\section{Introduction}
With the rapid advancement of Large Language Models (LLMs), Vision-Language Models (VLMs) like Shikra \cite{chen2023shikra} and MiniGPT-v2 \cite{chen2023minigptv2} have profoundly changed the landscape of Multi-modal Large Language Models (MLLMs). These models exhibit a remarkable ability to engage in fluent vision-language conversations with humans and have generated new state-of-the-art (SoTA) on multi-granularity vision-language tasks \cite{zhan2023mono3dvg}. LLaVA \cite{liu2023visual} has achieved great success in constructing instruction-following data to fine-tune the model, bringing new possibilities to the field of MLLMs. 
Despite these strides, it is crucial to note that the triumph of generalized MLLMs has not seamlessly extended to remote sensing (RS) vision-language tasks due to inherent differences between the natural and remote sensing domains.

Recently, there has been significant attention on remote sensing vision-language tasks \cite{10231134} of diverse granularity levels, including RS image captioning \cite{10138597}, RS visual question answering (VQA) \cite{10282946}, RS visual grounding \cite{10056343}, and UAV video captioning \cite{rs15082139}. Although efforts have been made to explore large vision-language models for remote sensing, it remains an emerging field with many challenges. 
The pioneering work, RSGPT \cite{rs15082139} is designed to address RS image captioning and VQA tasks using LLMs. However, it lacks the capability for multi-task conversation. RSGPT needs to train task-specific models on different datasets to solve tasks independently, which greatly limits its open-ended task capability.

\begin{figure}
  \centering
  \includegraphics[width=0.95\linewidth]{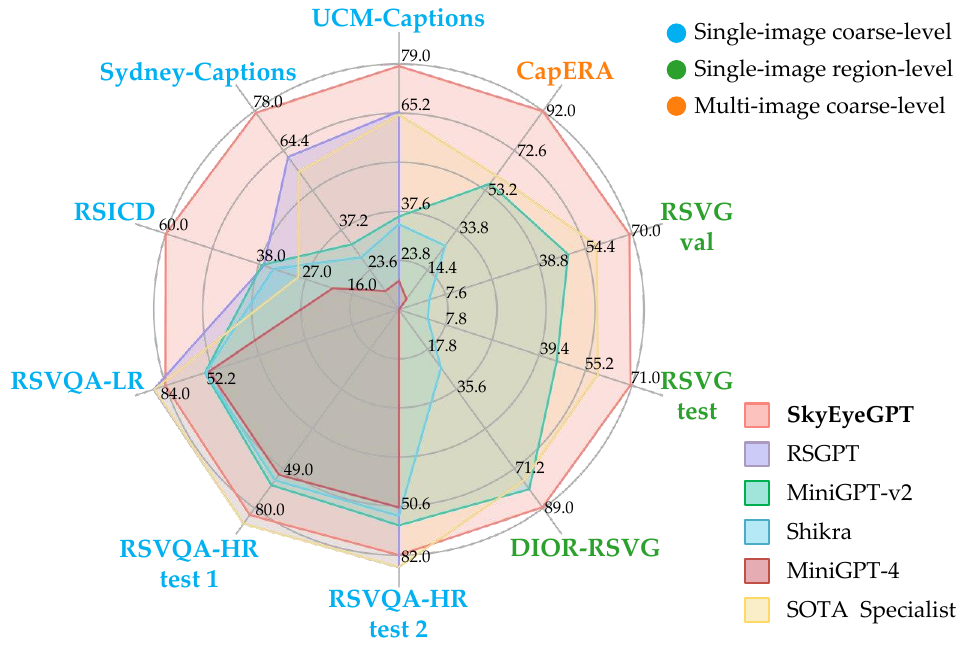}
  \caption{The performance of SkyEyeGPT on a broad range of RS vision-language tasks compared with existing models.}
  \label{fig:SkyEyeGPT-sota}
\end{figure}

\begin{figure*}[]
  \centering
  \includegraphics[width=0.95\linewidth]{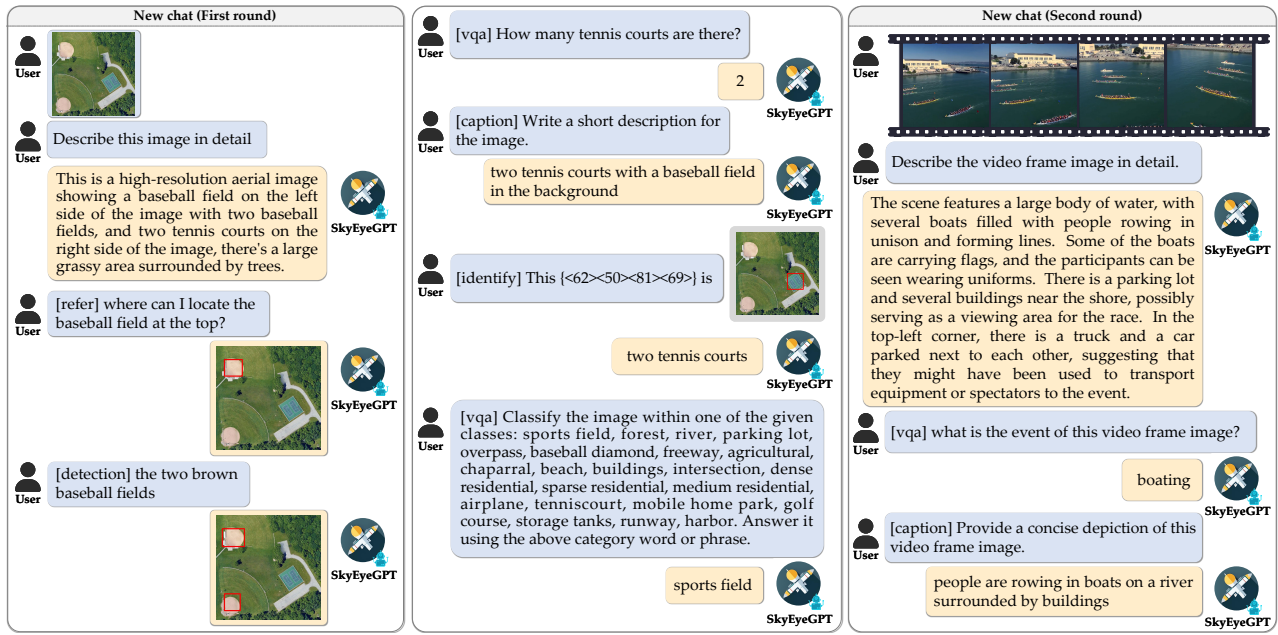}
    \captionof{figure}{\textbf{Remote Sensing Multimodal Conversational Interactions Facilitated by SkyEyeGPT.}
    The demonstration showcases SkyEyeGPT engaging in multi-task dialogues and completing various RS multi-modal tasks such as detailed image description, visual grounding, phrase grounding, VQA, image captioning, referring expression generation, scene classification, and UAV video captioning.
    }
  \label{fig:SkyEyeGPT_example}
\end{figure*}

Toward an open generalist framework that seamlessly combines the advantages of LLMs with remote sensing tasks, we introduce \textbf{SkyEyeGPT}, a unified model capable of handling open-ended RS vision-language tasks. The input and output of each task are represented in natural language, including bounding box coordinates. The SkyEyeGPT's architecture consists of a visual encoder, an alignment layer, and an LLM-based decoder for RS open-ended tasks. We do not design any extra encoder or external plugin modules, making SkyEyeGPT a unified and efficient model, and also simple to train and deploy.
Recent studies \cite{liu2023visual} have demonstrated the impressive results achieved by training MLLMs via instruction tuning, to connect LLMs and vision. 
Instruction tuning in the multi-modal domain of remote sensing is still underexplored. The challenge is the lack of large-scale RS multimodal instruction-following data.

To foster the research of RS VLMs, we meticulously curate an RS vision-language instruction-following dataset with 968k training samples, namely \textbf{SkyEye-968k}. Our instructions consist of the reorganization of public data and a few generated data. \emph{To guarantee correctness, data is manually verified and selected by our team members. We involve humans in the loop to ensure the high quality of the conversation instruction.} The SkyEye-968k is divided into single-task image-text instruction and multi-task conversation instruction. 
We set task-specific identifiers for different tasks to improve the ability of SkyEyeGPT on various specific tasks. To further explore multi-turn multi-task dialogue capabilities, we design a two-stage tuning that utilizes single-task and multi-task conversation instructions in two stages, respectively.

Experiments on 8 remote sensing vision-language datasets demonstrate SkyEyeGPT’s superiority, as shown in Figure \ref{fig:SkyEyeGPT-sota}. To further investigate whether SkyEyeGPT possesses good instruction-following ability, we compare it with MiniGPT-4, Shikra, MiniGPT-v2, and GPT-4V. 
We provide several real conversations with users and comparisons in Figure \ref{fig:SkyEyeGPT_example}, Figure \ref{fig:compared_results}, and Figure \ref{fig:GPT4_compare}.
Surprisingly, SkyEyeGPT, trained on our SkyEye-968k, shows results comparable to or even better than GPT-4V and can provide a more comprehensive and detailed understanding of remote sensing images.
To demonstrate the effectiveness of the simple SkyEyeGPT structure, we conduct extensive and adequate ablation studies with more detailed results in the supplementary materials.

Our contributions can be summarized as follows:
\begin{itemize}
\item {\it Unified RS vision-language instruction dataset, SkyEye-968k.} One challenge is the lack of instruction data for RS multi-modal large language model. We create high-quality instruction-following data, including single-task and multi-task conversation instruction.
\item {\it RS multi-modal large language model.} We develop SkyEyeGPT, which unifies RS vision-language tasks and breaks new ground in enabling the unified modeling of RS vision and LLM.
\item {\it Superior performance.} SkyEyeGPT achieves competitive performance on the image-level and region-level RS vision-language tasks. Specially, it has shown encouraging results in some tests, compared with GPT-4V.
\item {\it Open source SkyEyeGPT for real-world applications.} We release the following assets to the public community for applications in real-world scenarios: an online RS multi-modal chatbot, the model checkpoint, the instruction-following dataset, and the codebase.
\end{itemize}

\section{Related Work}
\label{relatedwork}

\subsection{Remote Sensing Vision-Language Tasks}
Recently, there has been significant attention on multi-modal tasks in remote sensing vision-language understanding \cite{10231134}. Traditional image-level tasks, such as RS image captioning and RS VQA, have made significant progress \cite{10282946}. Emerging region-level and spatio-temporal tasks, such as RSVG \cite{10056343} and UAV video captioning \cite{rs15082139}, have raised novel challenges and garnered increasing interest. Despite the availability of numerous state-of-the-art methods capable of performing these tasks \cite{xiong2022earthnets}, they are typically trained on a specific dataset to perform a specific task. This work primarily focuses on unifying the diverse RS vision-language tasks.

\subsection{LLMs for Vision-Language}
With the rise of advanced LLMs, ChatGPT \cite{Chatgpt}, LLaMA \cite{touvron2023llama}, GPT-4 \cite{Gpt4}, and Vicuna \cite{chiang2023vicuna} have shown remarkable abilities in various language tasks. BLIP-2 \cite{li2023blip} extends LLMs into the realm of multimodal by connecting the frozen LLM with a visual encoder via Q-Former. Some approaches employ the simplest linear layer as a mediator to link LLMs and visual encoders, achieving notable success, such as LLaVA \cite{liu2023visual}, MiniGPT-4 \cite{zhu2023minigpt}. Recent contributions from VisionLLM \cite{wang2023visionllm}, Shikra \cite{chen2023shikra}, and MiniGPT-v2 \cite{chen2023minigptv2} further substantiate that spatial coordinates in visual grounding tasks can be effectively handled in language form by LLM. 
These approaches showcase the potential and versatility of LLM for seamless integration of vision and language modalities. The application and research on generalized MLLMs in RS have been comparatively limited. RSGPT \cite{hu2023rsgpt} was the first attempt, but it could only handle coarse-grained tasks of image-text and doesn't support open-ended multi-tasks and multi-task conversations.

\subsection{Vision-Language Instruction Tuning}
The purpose of instruction tuning is to enhance the instruction following ability of the model. Drawing inspiration from LLMs in instruction tuning, LLaVA \cite{liu2023visual} fine-tunes the model based on synthetic multi-modal instruction-following data. Instruct-BLIP \cite{instructblip} collects a larger set of instruction data, resulting in improved performance for BLIP. These methods primarily focus on image-level coarse-grained tasks, and cannot effectively address fine-grained perception challenges.
Recent VisionLLM \cite{wang2023visionllm}, Shikra \cite{chen2023shikra}, and MiniGPT-v2 \cite{chen2023minigptv2} further utilize instruction-following data to tackle fine-grained visual perception tasks such as visual grounding, region caption, and object detection. These methods demonstrate the potential of instruction tuning strategies to mine the LLM's ability to understand and respond to multi-grained multi-modal instructions. 
Our method aims to provide a unified framework for handling open-ended RS vision-language tasks and develop multi-task conversational capability via instruction tuning.

% \section{Methodology}
\section{Method of SkyEyeGPT}
\label{sec:methods}

\begin{figure}
  \centering
  \includegraphics[width=0.99\linewidth]{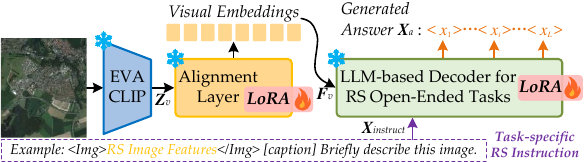}
    \caption{The overall framework of the proposed SkyEyeGPT.}
  \label{fig:ourSkyEyeGPT}
\end{figure}

\subsection{Overall Architecture}
\label{sec:model}

As depicted in Figure \ref{fig:ourSkyEyeGPT}, SkyEyeGPT consists of a visual encoder, an alignment layer, and an LLM-based decoder for RS open-ended tasks. \emph{More detailed comparisons of existing MLLMs are presented in the supplementary materials.}

\textbf{Visual Encoder.} 
The pre-trained vision transformer, EVA-CLIP \cite{Fang_2023_CVPR}, is employed as the visual encoder. The parameters are frozen during our training. 
Given an input RS image $\boldsymbol{I} \in \mathbb{R}^{H \times W \times3}$, $H$ and $W$ represent the height and width, respectively. Initially, the resolution of remote sensing images is standardized to 448×448. Subsequently, we apply the EVA model to segment the image into patches and extract image embeddings $\boldsymbol{Z}_{v} \in \mathbb{R}^{N \times D} $ from these patches, where $N$ is the number of patches and $D$ is the hidden dimension. The UAV video features are formed by the concatenation of features from multiple frame images.

\textbf{Alignment Layer.} 
We consider a linear layer to bridge the modality gap, aligning RS visual features from the visual encoder with the language features from advanced LLM. The input resolution is crucial for accurately understanding detailed RS image-text representations. However, the high resolution of $\text{448}\times \text{448}$ will generate an excessive number of patches $N$, which reduces the efficiency of processing contextual input in the LLM and is highly resource-demanding.
Therefore, we opt not to directly project the RS image embeddings into the linear layer. A simple yet effective method \cite{chen2023minigptv2} is adopted to directly concatenate four adjacent visual tokens to reduce the number of patches by four times. 
The linear layer converts the visual tokens $\boldsymbol{Z}_{v}^{'} \in \mathbb{R}^{\frac{N}{4} \times (4 \times D)} $ into embeddings $\boldsymbol{F}_{v} \in \mathbb{R}^{\frac{N}{4} \times d} $ in the language space, where $d$ is the hidden dimension size of LLM.

\textbf{LLM-based Decoder for RS Open-Ended Tasks.} 
We choose open-sourced LLaMA2-chat \cite{touvron2023llama2} as our language model, which is a decoder-only LLM. Our decoder takes a sequence of visual tokens $\boldsymbol{F}_{v}$ and language instructions as input, generating task-specific answers.
We acknowledge the existence of more sophisticated (but expensive) methods for connecting remote sensing images and language, such as Q-former in BLIP-2 \cite{li2023blip}, or other encoders like RemoteCLIP pre-trained on remote sensing data. We explore potentially more efficient or sophisticated architectures for SkyEyeGPT in ablation experiments.

\begin{table}
  \centering
\scalebox{0.87}{
\begin{tabular}{ccc}
\hline
\textbf{Task}   & \textbf{Data Source}  & \textbf{Samples}  \\ \hline
\multirow{5}{*}{\begin{tabular}[c]{@{}c@{}}Image\\ Captioning\end{tabular}}    & RSICD \cite{8240966}     &  43.7k   \\
& RSITMD \cite{9437331} & 21.5k   \\
& UCM-Captions \cite{7546397}    & 8.4k  \\
& Sydney-Captions \cite{7546397}  & 2.5k   \\
& NWPU-Captions \cite{9866055} & 126.0k  \\ \hline
\begin{tabular}[c]{@{}c@{}}Video\\ Captioning\end{tabular}  & CapERA \cite{rs15082139}  &  7.4k     \\ \hline
\multirow{4}{*}{\begin{tabular}[c]{@{}c@{}}Visual\\ Question\\ Answering\end{tabular}} & ERA-VQA (Ours)   & 1.5k  \\
& RSIVQA \cite{9444570}   & 17.7k   \\ 
& RSVQA-LR \cite{9088993} &  57.2k     \\
& RSVQA-HR \cite{9088993}  & 625.3k     \\ \hline
\multirow{2}{*}{\begin{tabular}[c]{@{}c@{}}Visual\\ Grounding\end{tabular}}  & RSVG \cite{3548316}  &  5.5k       \\
& DIOR-RSVG \cite{10056343} & 27.0k      \\ \hline
\begin{tabular}[c]{@{}c@{}}Phrase\\ Grounding\end{tabular}  & RSPG (Ours)   &  6.8k      \\ \hline
\multirow{4}{*}{\begin{tabular}[c]{@{}c@{}}Multi-task\\ Conversation \end{tabular}}  
& {DOTA-Conversa}$^{*}$ \text{(Ours)} & 1.4k  \\
& {DIOR-Conversa}$^{*}$ \text{(Ours)} & 14.7k  \\
& {UCM-Conversa}$^{*}$ \text{(Ours)} & 1.7k  \\
& {Sydney-Conversa}$^{*}$ \text{(Ours)} & 0.5k  \\ \hline
\end{tabular}
}
\caption{
Details on the training samples used for the RS vision-language instruction. The asterisk indicates that this data is only used in the second stage.}
 \label{datasets}
\end{table}

\subsection{Unified RS Vision-Language Instruction}
\label{sec:Instruction}
While acquiring instruction fine-tuning datasets in the general domain is straightforward, there are no equivalent datasets in the remote sensing domain. To address this gap, a unified RS vision-language instruction data, SkyEye-968k, is carefully planned and specifically tailored for the RS vision-language large model.
Our instruction data with 968k training samples consists of the reorganization of public data and a few generated data verified manually. Details are summarized in Table \ref{datasets}. We ensure that no images from the validation or test sets appear in the instructions, thus eliminating the risk of data leakage. The SkyEye-968k dataset is divided into two parts:

\textbf{Single-task Image-text Instruction.} 
\begin{itemize}
\item Captioning task. Specifically, we integrate five RS image captioning (RSICD, RSITMD, UCM-Captions, Sydney-Captions, and NWPU-Captions) and one UAV video captioning dataset (see Table \ref{datasets}).
\item VQA task. We integrate three public RS VQA datasets (RSIVQA, RSVQA-LR, and RSVQA-HR). The ERA-VQA dataset is generated based on the event recognition in aerial videos (ERA) dataset \cite{9295448}. Take frame images and questions about the event theme as input.
\item Grounding task. We integrate two public RS visual grounding datasets (RSVG and DIOR-RSVG). Following the method of generating object parsing and grounding instructions \cite{chen2023minigptv2}, we created an RS phrase grounding dataset (RSPG). With RS images and phrases as input, the output target bounding box can be either single or multiple.
\end{itemize}

\textbf{Multi-task Conversation Instruction.} 
Single-task instruction focuses only on high-quality aligned image-text data to improve SkyEyeGPT's performance on each specific task. After the first stage of tuning, when engaging in multiple rounds of conversations with the user, the model may struggle to handle subsequent tasks effectively as the context becomes more complex. 
To transform SkyEyeGPT into a proficient chatbot, we must focus on how to enhance its multi-task conversation capabilities, ensuring a good and seamless user experience. To tackle this challenge, we create the RS multi-task conversation instruction by mixing or reorganizing datasets from different tasks.

Specifically, we mix the corresponding captioning and VQA datasets to get UCM-Conversa and Sydney-Conversa instruction. Using the DIOR-RSVG dataset and DIOR dataset \cite{li2020object}, we construct DIOR-Conversa instruction which contains visual grounding, phrase grounding, and referring expression generation tasks. Similarly, we leverage RSIVQA and the DOTA object detection dataset \cite{xia2018dota} to build a conversation instruction, DOTA-Conversa, that includes VQA and phrase grounding tasks.
To guarantee correctness, data is manually verified and selected by our team members. We involve humans in the loop to ensure the high quality of the instruction.

\begin{table}
    \centering
    \begin{tabular}{l} 
        \includegraphics[width=0.95\linewidth]{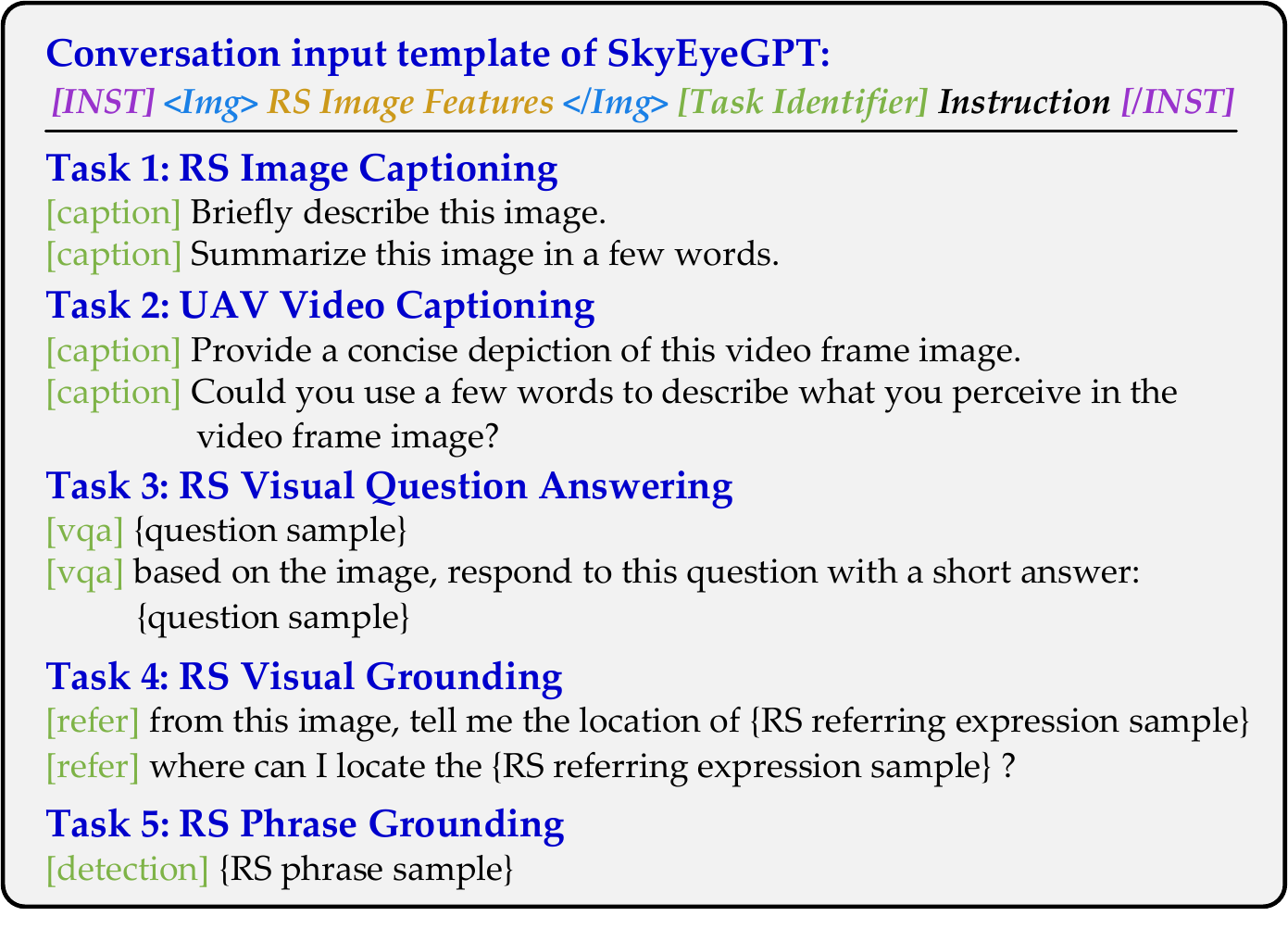} \\ 
    \end{tabular} 
    \caption{Conversation input template and instruction examples (randomly chosen examples) for each task.} 
    \label{tab:instruction_example} 
\end{table}

\begin{table*}[h]
\centering
\scalebox{0.9}{
\begin{tabular}{lcccccccc}
\hline
 \textbf{Method} & Open-Ended & BLEU-1 & BLEU-2 & BLEU-3 & BLEU-4 & METEOR & ROUGE\_L & CIDEr  \\ \hline
\multicolumn{9}{l}{\textit{\textbf{Specialist Models}: representative or SoTA methods with results reported in the literature}} \\
SAA \cite{8931249} &\ding{55}&79.62 &74.01 &69.09 &64.77 &38.59& 69.42 &294.51   \\
Post-processing \cite{10138597} &\ding{55}& 79.73  &  72.98 & 67.44 & 62.62 & 40.80 & 74.06 & {309.64}   \\  \hline
\multicolumn{9}{l}{\textit{\textbf{Generalist Models}: results of our own experimental runs (except RSGPT)}} \\
MiniGPT-4 \cite{zhu2023minigpt}&\ding{51}& 30.90  & 27.55  & 22.23  & 18.10  & 33.36 & 41.37    & 0.03  \\
Shikra \cite{chen2023shikra}&\ding{51}& 81.16&58.94&43.26&33.98 &32.56 & 56.73 & 56.69 \\
MiniGPT-v2 \cite{chen2023minigptv2}&\ding{51}& 81.10  & 60.27  & 45.10  & 36.16  & 32.41  & 56.57    & 60.66  \\
RSGPT \cite{hu2023rsgpt}&\ding{55} & 86.12  & 79.14  & 72.31  & 65.74  & 42.21  & 78.34    & \textbf{333.23} \\ 
% \rowcolor{blue!7} 
\rowcolor{violet!10}
$\text{SkyEyeGPT}_{single}$ &\ding{55} &\textbf{92.03} & \underline{85.66} & \underline{80.63} &\underline{76.76} &\textbf{46.24} & \textbf{80.10} &220.99 \\
\rowcolor{violet!10}
$\text{SkyEyeGPT}_{one-stage}$ &\ding{51} &90.58 & 83.97 & 78.52 &74.41 &\underline{45.10} & 77.41 &220.36 \\
\rowcolor{violet!10}
SkyEyeGPT  &\ding{51} & \underline{90.71}  & \textbf{85.69}  & \textbf{81.56}  & \textbf{78.41}  & \textbf{46.24}  & \underline{79.49}    & \underline{236.75} \\ \hline
\end{tabular}
}
\caption{Comparisons with Generalist and Specialist models on \textbf{UCM-captions dataset} for RS image captioning task.}
\label{UCMcaption}
\end{table*}

\begin{table*}[h]
\centering
\scalebox{0.9}{
\begin{tabular}{lcccccccc}
\hline
 \textbf{Method}  & Open-Ended & BLEU-1 & BLEU-2 & \multicolumn{1}{c}{BLEU-3} & \multicolumn{1}{c}{BLEU-4} & METEOR & ROUGE\_L & CIDEr \\ \hline
\multicolumn{9}{l}{\textit{\textbf{Specialist Models}: representative or SoTA methods with results reported in the literature}} \\
 CapERA \cite{rs15082139} &\ding{55}& 50.43  & 37.26  & 29.24   & 22.90  & 21.16  & 43.90   & 60.42 \\ \hline
 \multicolumn{9}{l}{\textit{\textbf{Generalist Models}: results of our own experimental runs}} \\
 MiniGPT-4 \cite{zhu2023minigpt}&\ding{51}& 25.20  & 22.98 & 18.57 & 14.98 & 30.40 & 33.71& 0.05  \\
 Shikra \cite{chen2023shikra}&\ding{51}& 79.11 & 60.57 & 46.29& 37.29& 32.06 & 56.47 & 26.16\\
 MiniGPT-v2 \cite{chen2023minigptv2} &\ding{51}& \textbf{81.82}  & 63.62  & 49.31   & 39.93  & 32.61  & 57.25   & 56.47 \\
 % \rowcolor{blue!7} 
 \rowcolor{violet!10}
$\text{SkyEyeGPT}_{single}$ &\ding{55}& \underline{80.40}&\underline{68.52}  & \underline{58.59} &\underline{51.49} &\textbf{34.08} & \underline{63.31} & \underline{90.92}\\
\rowcolor{violet!10}
$\text{SkyEyeGPT}_{one-stage}$ &\ding{51} & 75.74& 64.64& 55.62 &49.20 &31.82 &61.49  &85.11 \\
 \rowcolor{violet!10}
SkyEyeGPT & \ding{51} & 80.13  & \textbf{69.04} & \textbf{59.56} & \textbf{52.74} & \underline{33.97} & \textbf{63.67} & \textbf{91.90} \\ \hline
\end{tabular}
}
\caption{Comparisons with Generalist and Specialist models on \textbf{CapERA dataset} for aerial video captioning task.}
\label{videocaption}
\end{table*}

\subsection{Instruction Tuning}
\label{sec:Instruction_train}
The model is trained to follow a series of task-specific instructions on the RS multi-modal instruction-following data. The input template and instruction examples for SkyeyeGPT are illustrated in Table \ref{tab:instruction_example}. To achieve an effective SkyEyeGPT, we design a two-stage instruction tuning approach.

\textbf{Input and Output Template.}
We build a variety of task inputs, following the conversation input template in Table \ref{tab:instruction_example}. We introduce task-specific identifiers, such as \textquotedblleft \textcolor[RGB]{126,179,71}{[caption]}, \textcolor[RGB]{126,179,71}{[vqa]}, \textcolor[RGB]{126,179,71}{[refer]}\textquotedblright. This design achieves the unification of RS vision-language tasks while allowing the model to flexibly produce task-specific outputs. The answer or response, \textit{i.e.}, the model output, follows after the \textcolor[RGB]{153,51,204}{\textit{[/INST]}}.
The input or output of region-level tasks requires bounding boxes of objects. We represent the coordinates in the natural language form $\{$\textless $x_1$\textgreater \textless $y_1$\textgreater \textless $x_2$\textgreater \textless $y_2$\textgreater$\}$. Specifically, ($x_1$, $y_1$) and ($x_2$, $y_2$) denote the coordinates of the top-left and bottom-right corners of the box, respectively. The coordinate values are normalized, multiplied by 100, and rounded to integers.

\textbf{Stage 1: Remote Sensing Image-Text Alignment.}
This stage trains the model using the single-task image-text instruction. This helps SkyEyeGPT build remote sensing fine-grained knowledge of multi-tasking.
Treat each sample as a single-round conversation $X_{c} = (\boldsymbol{X}_{instruct}, \boldsymbol{X}_{a})$. For the given \textcolor[RGB]{204,153,26}{RS image features} $\boldsymbol{F}_{v}$, connect it with the instruction tokens $\boldsymbol{X}_{instruct}$ from the text modality. This concatenated input is then fed into the LLM. SkyEyeGPT generates the answer $\boldsymbol{X}_{a}$ with a length of $L$. Maximizing the likelihood function that is defined as follows:

\begin{equation}
\label{loss}
\begin{split}
    \mathcal{L} & = \text{log}P \left (\boldsymbol{X}_{a} \mid \boldsymbol{F}_{v},\boldsymbol{X}_{instruct} ; \theta \right )   \\
                & = \sum_{i=1}^{L} \text{log}P \left ( x_{i} \mid \boldsymbol{F}_{v},\boldsymbol{X}_{instruct},\boldsymbol{X}_{a,<i} ; \theta \right ),
\end{split}
\end{equation}
where $P$ and $\theta$ are the conditional probability and the trainable parameters, and $\boldsymbol{X}_{a,<i}$ is the answer tokens preceding the current prediction tokens $x_{i}$.

\textbf{Stage 2: Multi-task Conversation Fine-tuning.} 
This stage uses the multi-task conversation instruction to better answer questions for multiple rounds and multiple tasks, enabling SkyEyeGPT to generate more natural and convincing outputs in multi-task conversations. 
The multi-task conversation is represented as a list $X_{c} = (\boldsymbol{X}_{instruct}^{1}, \boldsymbol{X}_{a}^{1},... ,$ $ \boldsymbol{X}_{instruct}^{n}, \boldsymbol{X}_{a}^{n})$, where $\boldsymbol{X}_{instruct}^{n}$ is the instruction for $n$-th turn. Similarly, the objective function is as follows:

\begin{equation}
\label{loss2}
\begin{split}
    \mathcal{L} & = \text{log}P \left (\boldsymbol{X}_{a} \mid \boldsymbol{F}_{v},\boldsymbol{X}_{instruct} ; \theta \right )   \\
                & = \sum_{i=1}^{L} \text{log}P \left ( x_{i} \mid \boldsymbol{F}_{v},\boldsymbol{X}_{instruct,<i},\boldsymbol{X}_{a,<i} ; \theta \right ),
\end{split}
\end{equation}
where $\boldsymbol{X}_{instruct,<i}$ is the instruction tokens in all turns before the current prediction tokens $x_{i}$. Therefore, the instructions and answers from previous rounds serve as references for the current task's response.

In the above two stages, we employ the Low-Rank Adaptation (LoRA) method to fine-tune the alignment layer and LLM, as shown in Figure \ref{fig:ourSkyEyeGPT}. This approach can fine-tune the model with limited resources and promote alignment between the two modalities of remote sensing vision and language.

\section{Experiments}
\label{sec:experiments}

\subsection{Experimental Details}
The parameters of the linear layer and LLM are initialized from MiniGPT-v2’s checkpoint \cite{chen2023minigptv2}. In the first stage, we finetune our SkyEyeGPT end-to-end for 35 epochs on the single-task image-text instruction. In the second stage, we add the multi-task conversation instruction and reduced the sampling ratio of single-task instruction to train 5 epochs of SkyEyeGPT.
Our setting is the same for the two training stages. The AdamW is used as the optimizer. We set the batch size to 1 with $\text{10}^{-\text{5}}$ learning rate and a cosine learning rate scheduler. To control overfitting, we apply a weight decay of 0.05. The rank in LoRA is 64. All training is conducted on four NVIDIA 3090 GPUs.

\subsection{Remote Sensing Multi-modal Chatbot} 
\begin{figure*}[h]
  \centering
  \includegraphics[width=0.95\linewidth]{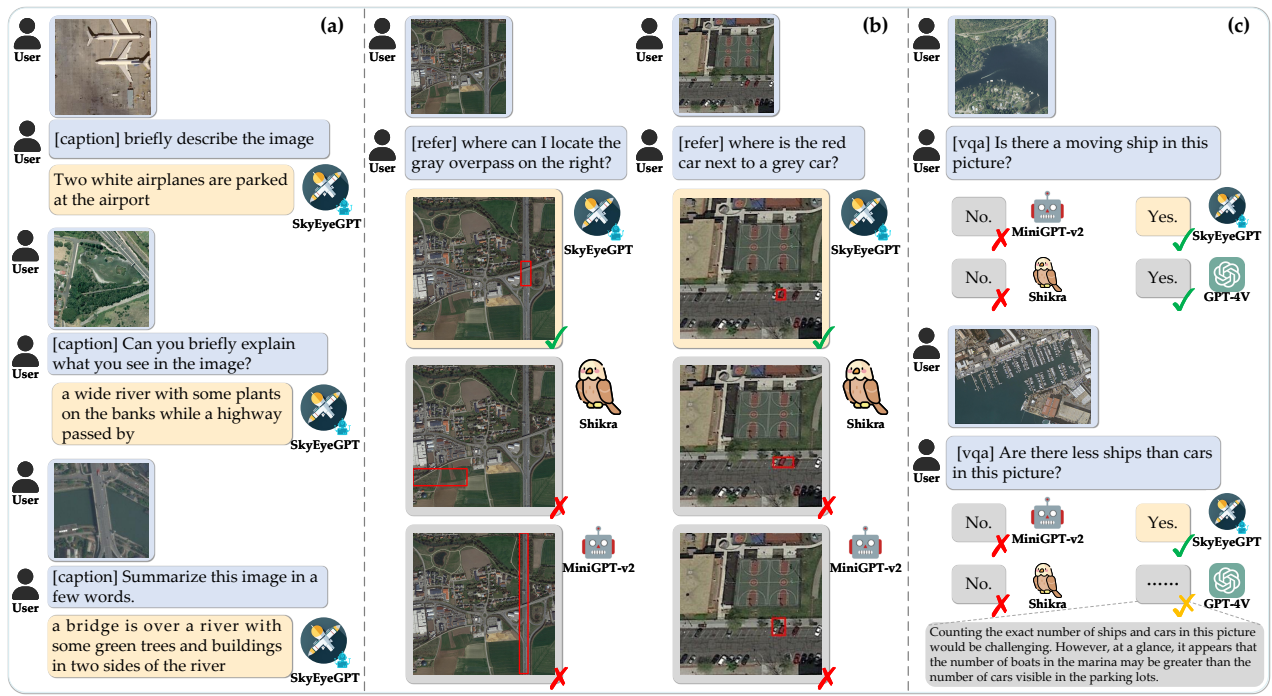}
  \caption{
 Some testing samples of captioning, grounding, and VQA. SkyEyeGPT has demonstrated impressive performance.
  }
  \label{fig:compared_results}
\end{figure*}

As shown in Figure \ref{fig:SkyEyeGPT_example}, we have developed a demonstration of a remote sensing multi-modal chatbot, showcasing the vision-language understanding and conversational capabilities of SkyEyeGPT.
We also provide several real conversations with users or comparisons in Figure \ref{fig:compared_results}, Figure \ref{fig:GPT4_compare}, and supplementary materials. Surprisingly, SkyEyeGPT, trained on our RS instruction-following dataset, demonstrates results comparable to or even better than GPT-4V.

\subsection{Main Results}
We conduct experiments on four representative tasks: RS image captioning, UAV video captioning, RS visual question answering, and RS visual grounding. Specialist models are designed for specific tasks, and we only report a few latest SoTA methods. Generalist models can perform various vision-language tasks.

% \subsection{RS Image Captioning}
\textbf{RS Captioning:} 
Image caption is crucial for assessing the quality of the alignment between RS vision and language. For this task, the model generates a description based on the user-input RS image and instruction. 
As shown in Tables \ref{UCMcaption} and \ref{videocaption}, we achieved the best performance in most of the metrics except for CIDEr on the UCM-caption and achieved SoTA results on the CapERA dataset for aerial video captioning. 
MiniGPT-4 generates longer captions with rich details, so it is difficult for existing captioning evaluation metrics to provide accurate evaluation, especially CIDEr. RSGPT achieves high CIDEr scores due to its fine-tuning on each dataset to produce results with similar lengths. \emph{We employ a novel ChatGPT-based evaluation method in the supplementary material.}

\begin{figure*}
  \centering
  \includegraphics[width=0.9\linewidth]{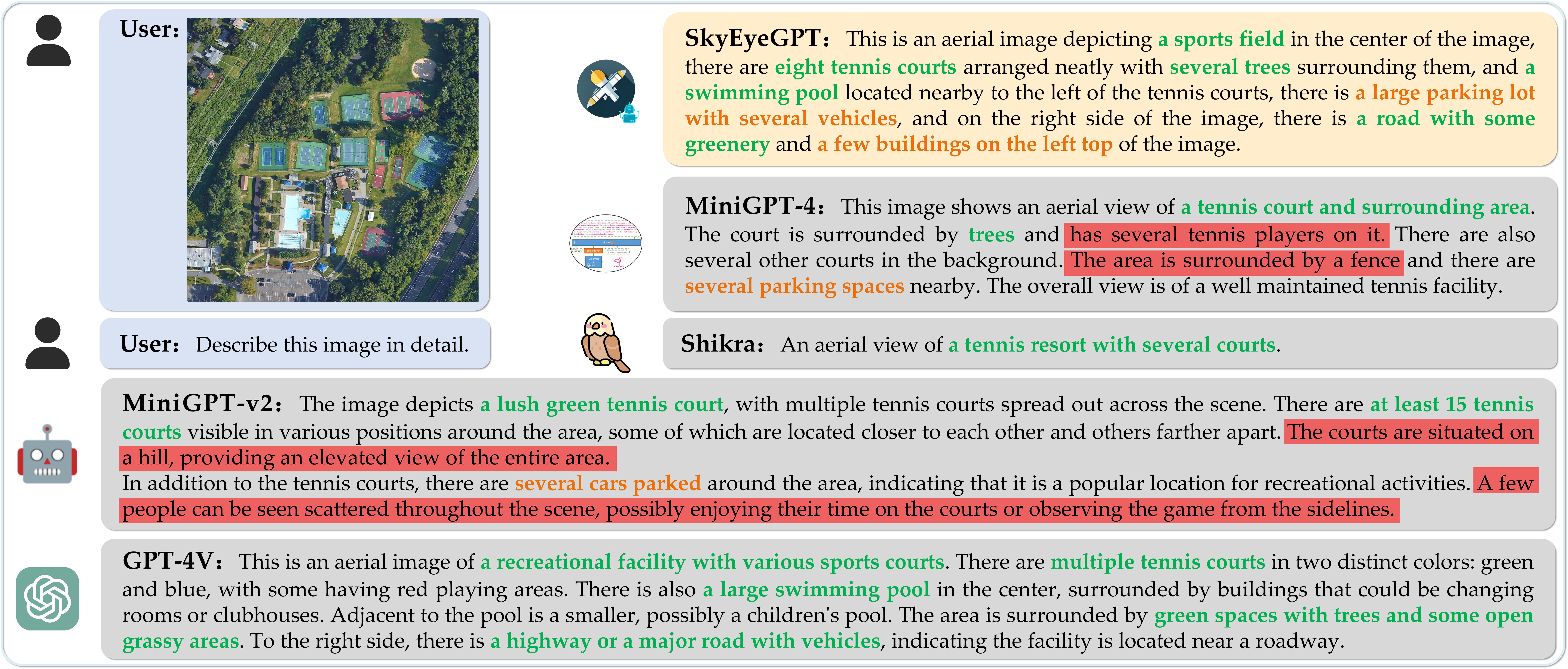}
  \caption{
  Detailed description results on RS images with complex scenes demonstrate the comparable and encouraging remote sensing visual understanding capability of SkyEyeGPT compared to GPT-4V.
  }
  \label{fig:GPT4_compare}
\end{figure*}

Figures \ref{fig:compared_results} (a) and \ref{fig:GPT4_compare} show some qualitative results for image caption and comparison for the detailed description, respectively.
% Figure \ref{fig:GPT4_compare} presents the comparison for the detailed description. 
The first sentence is an overview of the images, while MiniGPT-4, Shikra, and MiniGPT-v2 are limited to tennis courts only, whereas SkyEyeGPT and GPT-4V describe it better as a sports field or court. Both MiniGPT-4 and MiniGPT-v2 exhibit errors in descriptions, whereas Shikra excels in referential dialogue and is not suitable for detailed description. GPT-4V performs admirably, but it ignores the buildings at the top left of the image and the parking lot at the bottom, which SkyEyeGPT accurately describes.

% \subsection{RS Visual Grounding}
\textbf{RS Visual Grounding:}
The model receives an RS image and a referring expression, then outputs the bounding box referring to the target object. 
Quantitative results for the test set of DIOR-RSVG and the validation and test sets of RSVG are provided in Table \ref{result-rsvg}. 
The RSVG dataset requires the model to have a stronger numerical geospatial relations understanding. Our testing has shown that Shikra has poor robustness, performing poorly outside of the training set domain, and cannot be used on the more challenging RSVG dataset.
SkyEyeGPT outperforms the SoTA specialist models by about 10\%.
Figure \ref{fig:compared_results} (b) displays test examples from NWPU and DOTA images, indicating that SkyEyeGPT has good robustness and precise localization ability for small objects.

\begin{table}[]
\centering
\scalebox{0.85}{
\begin{tabular}{lccc}
\hline
\multirow{2}{*}{\textbf{Method}} & \multicolumn{2}{c}{\textbf{RSVG}} & \textbf{DIOR-RSVG}  \\  \cmidrule(r){2-3}\cmidrule(r){4-4}
 & val       & test     & test     \\ \hline
\multicolumn{4}{l}{\textit{\textbf{Specialist Models}: representative or SoTA methods with results}} \\
% representative or SoTA methods with results reported in the literature
 FAOA \cite{yang2019fast} & 30.06     & 30.15     & 67.21     \\
ReSC \cite{yang2020improving} & 53.96      & 51.18     & 72.71     \\
LBYL-Net \cite{Huang_2021_CVPR}& 31.64     & 32.19     & 73.78     \\
GeoVG \cite{3548316} & 58.20    & 59.40     & /      \\
MGVLF \cite{10056343} & /         & /     & 76.78     \\ \hline
\multicolumn{4}{l}{\textit{\textbf{Generalist Models}: results of our own experimental runs}} \\
Shikra \cite{chen2023shikra} & 2.16  & 1.87  & 26.08 \\
MiniGPT-v2 \cite{chen2023minigptv2} & 48.63  & 45.48    & 80.47     \\
% \rowcolor{blue!7} 
\rowcolor{violet!10}
$\text{SkyEyeGPT}_{single}$ &\textbf{70.02} & \underline{69.68} & \underline{87.64} \\
\rowcolor{violet!10}
$\text{SkyEyeGPT}_{one-stage}$  & {67.49}  & 66.98  & 86.24 \\
\rowcolor{violet!10}
SkyEyeGPT & \underline{69.19} &\textbf{70.50} &\textbf{88.59} \\ \hline
\end{tabular}
}
 \caption{Comparisons with Generalist and Specialist models on \textbf{RSVG} and \textbf{DIOR-RSVG datasets} for RS visual grounding task.}
\label{result-rsvg}
\end{table}

% \subsection{RS Visual Question Answering}
\textbf{RS VQA:} 
Table \ref{RSVQA} shows the results of our RSVQA evaluation. 
% We report top-1 accuracy for each subclass and average accuracy.
RSGPT separately fine-tuned on the RSVQA dataset yields high performance. Our average accuracy is 8\% lower than RSGPT. Moreover, the images in RSVQA belong to satellite imagery in our SkyEye-968k, while other RS images belong to aerial images. The image modality difference of RSVQA leads to performance loss. 
Addressing the modality difference problem of RS images from different sources is a key focus of our future work.
Figure \ref{fig:compared_results} (c) presents test samples from DOTA images, where SkyEyeGPT and GPT-4V have impressive performance.
We provide more detailed results in the supplementary materials, including the results for Sydney-caption, RSICD, and RSVQA-HR test set 1.

\begin{table*}[t]
\centering
\scalebox{0.82}{
\begin{tabular}{lcccccccc}
\hline
\multirow{2}{*}{\textbf{Method}}  & \multirow{2}{*}{\begin{tabular}[c]{@{}l@{}}Open-\\ Ended\end{tabular}}  & \multicolumn{4}{c}{\textbf{RSVQA-LR Test Set}}    & \multicolumn{3}{c}{\textbf{RSVQA-HR Test Set 2}} \\ \cmidrule(r){3-6}\cmidrule(r){7-9}
& & Presence & Comparison & Rural/Urban & Average Acc & Presence & Comparison & Average Acc \\ \hline
\multicolumn{9}{l}{\textit{\textbf{Specialist Models}: representative or SoTA methods with results reported in the literature}} \\
EasyToHard \cite{9771224}  & \ding{55}&  90.66  & 87.49   & 91.67   & 89.94 &  87.97  & 87.68  & 87.83  \\
SHRNet  \cite{10018408}  & \ding{55} &91.03 & 90.48 & {94.00} & 91.84 & 89.81  &  89.44 & 89.63 \\ \hline
\multicolumn{9}{l}{\textit{\textbf{Generalist Models}: results of our own experimental runs (except RSGPT)}} \\
MiniGPT-4 \cite{zhu2023minigpt}&\ding{51}&43.86&57.55&62.00&54.47&50.43&52.60&51.52\\ 
Shikra \cite{chen2023shikra} &\ding{51}&46.47&60.31&63.62&56.80&57.28&56.63&57.00\\ 
MiniGPT-v2 \cite{chen2023minigptv2} &\ding{51} &49.85& 63.09& 59.00 & 57.31 & 66.34 & 59.40 & 62.87\\
RSGPT \cite{hu2023rsgpt}&\ding{55} & \textbf{91.17}  &  \textbf{91.70}  & \textbf{94.00}  & \textbf{92.29} & \textbf{89.87}  & \textbf{89.68}  & \textbf{89.78}  \\
 % \rowcolor{blue!7} 
\rowcolor{violet!10}
$\text{SkyEyeGPT}_{single}$ &\ding{55}& \underline{90.23} & \underline{90.46} &\underline{84.00}&\underline{88.23} & \underline{87.50}& \underline{86.24}& \underline{86.87} \\
\rowcolor{violet!10}
$\text{SkyEyeGPT}_{one-stage}$ &\ding{51} & 87.20  & 88.46 & 68.00& 81.22 & 80.00 & 80.26 & 80.13  \\
\rowcolor{violet!10}
SkyEyeGPT &\ding{51} & {88.93} & {88.63} &{75.00}  &{84.19} & {83.50}& {80.28}&{81.89} \\ \hline
\end{tabular}
}
\caption{\textbf{Results on RSVQA-LR test set and RSVQA-HR test set 2.} Comparison of VQA results by Generalist and Specialist baselines.}
\label{RSVQA}
\end{table*}

\subsection{Ablation Studies}
\label{sec:ablation}
In this section, we analyze the impact of key components and hyperparameters of SkyEyeGPT in detail. 
To explore the impact of SkyEyeGPT's multi-task learning and two-stage instruction tuning approach, we develop two variants ($\text{SkyEyeGPT}_{single}$ and $\text{SkyEyeGPT}_{one-stage}$) to compare single-task tuning and one-stage instruction tuning. ${single}$ indicates that the model is trained separately on each task and lacks open-ended task ability. 
As shown in Tables \ref{UCMcaption}-\ref{RSVQA}, except for the VQA task where the model without open-ended task ability is significantly better than SkyEyeGPT, their performance is comparable in other tasks, indicating a balance between SkyEyeGPT's accuracy and generalization.
The model trained with the multi-task conversation instruction in the second stage can significantly improve performance on various tasks.
To demonstrate that our alignment layer is sufficient to align RS visual and textual features, we design three variants: (a) w/o Linear Layer, (b) + Multiple Linear Layers, and (c) + Q-Former. 
We also compare the impact with and without task identifiers. We set different ranks in LoRA to explore the impact on the results. We provide more detailed ablation experiments and results in the supplementary material.

\section{Conclusion}
\label{sec:conclusion}
In this work, we introduce SkyEyeGPT, a unified open MLLM tailored specifically for remote sensing. 
We construct an RS multi-modal instruction-following dataset, including single-task and multi-task conversation instruction. We design a two-stage tuning method to develop the model's multi-task and multi-round conversational ability. Task-specific identifiers are set to facilitate a unified treatment of open-ended tasks.
The effectiveness and superiority of SkyEyeGPT are validated on a range of different granularity tasks. SkyEyeGPT achieves the new SoTA accuracy on many tasks and provides an exceptional remote sensing multi-modal chatting experience.
This work represents a significant advancement in the remote sensing multi-modal domain, offering a versatile and high-performing solution for open-ended tasks in a unified framework via LLM.

\section*{Appendices}

In the supplementary material, we provide detailed structural comparisons of existing MLLMs, instructions of various remote sensing multi-modal tasks, more quantitative results, more ablation studies of SkyEyeGPT, and more qualitative results.

\subsection*{A Architecture Comparison of MLLMs}

\begin{table*}[b]
\centering
\scalebox{0.95}{
\begin{tabularx}{\linewidth}{lcccY}
\hline
\textbf{Model} & \textbf{Visual encoder} & \textbf{Connection of multimodal} & \textbf{LLM} & \textbf{Non-frozen Components} \\ \hline
% VisionLLM & ResNet or InternImage-H &           &  Alpac&LoRA\\
MiniGPT-4 & ViT-G/14 (EVA-CLIP) & a Q-Former + a linear layer &  Vicuna   & linear layer \\
\multirow{2}{*}{LLaVA}  & \multirow{2}{*}{ViT-L/14 (CLIP)}  & \multirow{2}{*}{a linear layer}  & \multirow{2}{*}{LLaMA}  & \multirow{2}{*}{\begin{tabular}[c]{@{}c@{}}stage1: linear layer\\ stage2: linear layer \& LLM\end{tabular}} \\
  &  &  &  & \\
% LLaVA  & ViT-L/14 (CLIP) & a linear layer &LLaMA  &stage1: linear layer \ \ \ \ \ \ \ \ \ \ \ \  \  stage2: linear layer \& LLM \\
Shikra & ViT-L/14 (CLIP) & a fully connected layer  & Vicuna-7/13B &  fully connected layer \& LLM \\
MiniGPT-v2 & ViT (EVA-CLIP)  & a linear layer & LLaMA2 &linear layer \& LLM \\
u-LLaVA    & ViT-L/14 (CLIP) & a linear layer & LLaMA2 &linear layer \& LLM \\
% Monkey  & Vit-BigHuge  &  resampler \cite{alayrac2022flamingo}  & Qwen-7B & resampler \& LLM \\
RSGPT  & ViT-G (EVA) & a Q-Former + a linear layer & Vicuna-7/13B & Q-Former \& linear layer  \\
\rowcolor{violet!10}
\textbf{SkyEyeGPT} & ViT-G (EVA-CLIP) & a linear layer & LLaMA2 & linear layer \& LLM (LoRA) \\ \hline
\end{tabularx}
}
\caption{Detailed architecture comparison of existing SoTA MLLMs.}
\label{compa_MLLM}
\end{table*}

We extract the description of the structure of existing SoTA visual-language multi-modal large language models from the original literature, as follows:

\textbf{MiniGPT-4} \cite{zhu2023minigpt}: 
It utilizes an advanced large language model (LLM), Vicuna, which is built upon LLaMA. In terms of visual perception, we employ the same pre-trained vision components of BLIP-2 that consist of a ViT-G/14 from EVA-CLIP and a Q-Former network. We target to bridge the gap between the visual encoder and LLM using a linear projection layer.

\textbf{LLaVA} \cite{liu2023visual}: 
We choose LLaMA as our LLM. For an input image, we consider the pre-trained CLIP visual encoder ViT-L/14, which provides the visual feature. We consider a simple linear layer to connect image features into the word embedding space.

\textbf{Shikra} \cite{chen2023shikra}: 
We selected the pre-trained ViT-L/14 of CLIP as visual encoder and Vicuna-7/13B as our LLM. We use one fully connected layer to map the ViT’s output embedding V to V' for modal alignment and correct input dimension of LLM.

\textbf{MiniGPT-v2} \cite{chen2023minigptv2}: 
It consists of three components: a visual backbone, a linear projection layer, and a large language model. MiniGPT-v2 adapts the EVA ViT as our visual backbone model backbone. MiniGPT-v2 adopts the open-sourced LLaMA2-chat (7B) as the language model backbone.

\textbf{u-LLaVA} \cite{xu2023u}: 
To align representations among different modalities, the projector-based structure is adopted in this work: the pre-trained CLIP ViT-L/14 and a visual projector are combined to encode image inputs, while the LLaMA2 is employed as the cognitive module. The vision projector for representation alignment and the hidden state projector for segmentation are two MLPs with channels of [1024, 4096] and [256, 4096, 4096].

\textbf{RSGPT} \cite{hu2023rsgpt}: 
Off-the-shelf frozen pre-trained image encoders (EVA-G) and large language models (vicuna7b, vicuna13b) form the foundation of the model. Following InstructBLIP, an instruction-aware Query Transformer (Q-Former) is inserted between them to enhance the alignment representation of visual features and textual features. Furthermore, a linear layer is introduced to project the output features of the Q-Former into the input features of LLM.

We have sorted out the key structures and summarized them in Table \ref{compa_MLLM}. All models employ ViT as their visual encoder, and the pre-trained models are sourced from either CLIP \cite{radford2021learning} or EVA \cite{Fang_2023_CVPR}. There are two types of connection between the vision feature and LLM, one follows InstructBLIP \cite{instructblip} which utilizes a Q-former and a linear layer, and the other employs only a linear layer. The LLM used in these models is chosen from the current most advanced open-source LLM. The visual encoder is all frozen during training, and the non-frozen components are mainly divided into linear layers and LLM.

\subsection*{B Instructions}

\textbf{Instructions for RS image captioning.}
The list of instructions for RS image captioning which briefly describes the RS image content is shown in Table \ref{tab:instruction_caption}. They present the same meaning with natural language variance.

\begin{table}[h]
    \centering
    \begin{tabular}{l} 
        \includegraphics[width=0.98\linewidth]{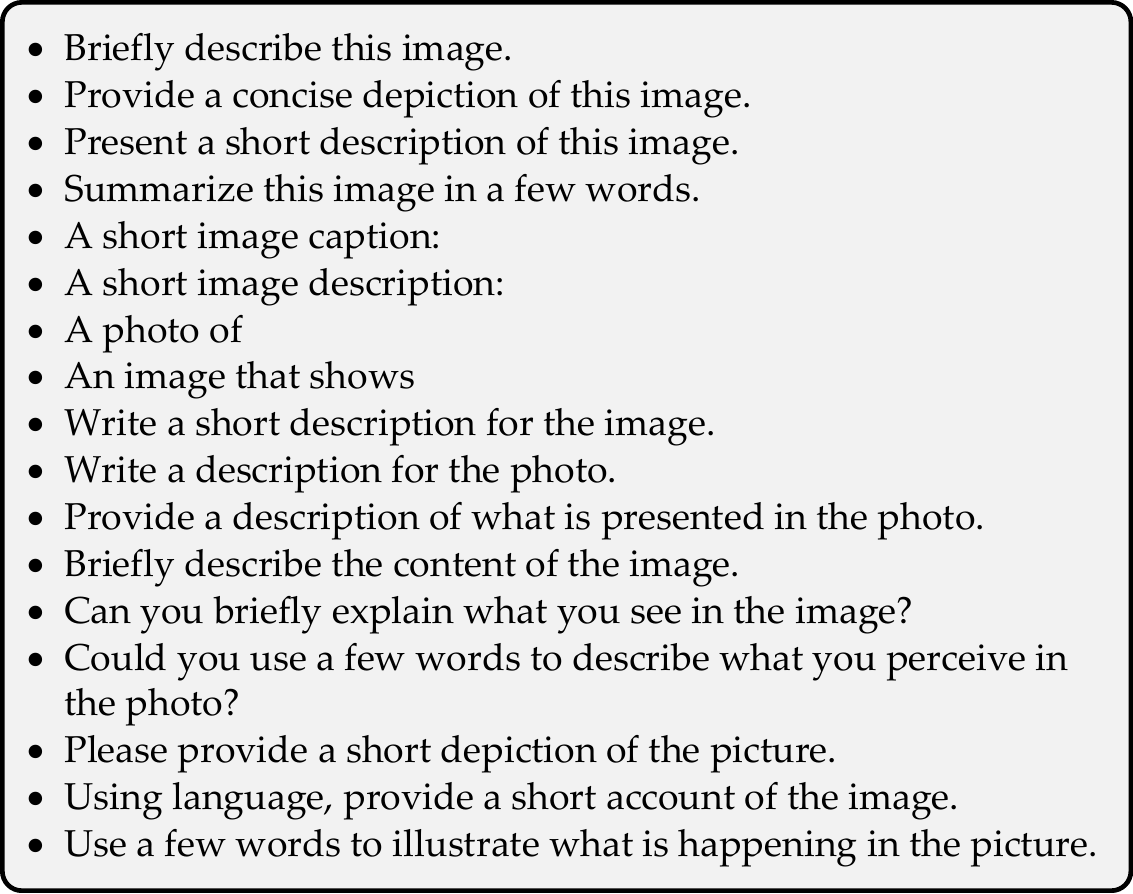} \\ 
    \end{tabular} 
    \caption{The list of instructions for RS image captioning.} 
    \label{tab:instruction_caption} 
\end{table}

\textbf{Instructions for RS visual grounding.}
The list of instructions for RS visual grounding which localizes the spatial location of the RS object is shown in Table \ref{tab:instruction_rsvg}. They present the same meaning with natural language variance.

\begin{table}[h]
    \centering
    \begin{tabular}{l} 
        \includegraphics[width=0.98\linewidth]{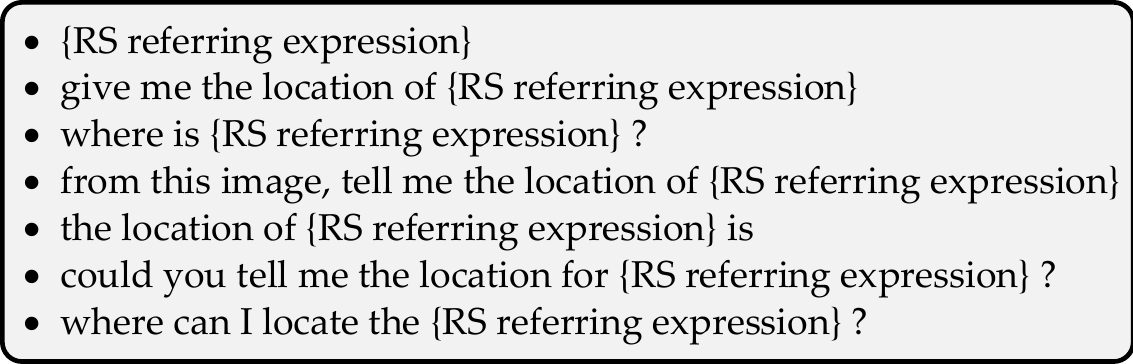} \\ 
    \end{tabular} 
    \caption{The list of instructions for RS visual grounding.} 
    \label{tab:instruction_rsvg} 
\end{table}

\textbf{Instructions for RS VQA.}
The list of instructions for RS VQA is shown in Table \ref{tab:instruction_rsvqa}. 

\begin{table}[h]
    \centering
    \begin{tabular}{l} 
        \includegraphics[width=0.98\linewidth]{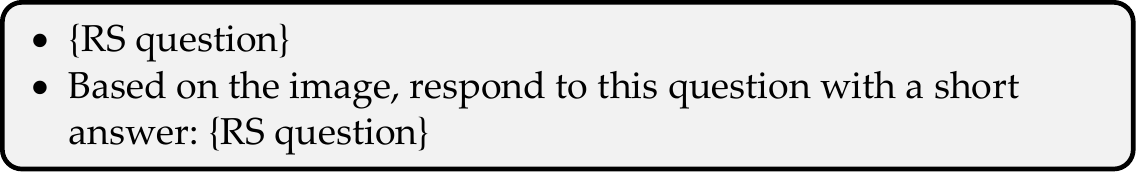} \\ 
    \end{tabular} 
    \caption{The list of instructions for RS VQA.} 
    \label{tab:instruction_rsvqa} 
\end{table}

\begin{table*}[b]
\centering
\scalebox{0.93}{
\begin{tabular}{lcccc}
\hline
\multirow{2}{*}{\textbf{Method}}  & \multirow{2}{*}{\begin{tabular}[c]{@{}l@{}}Open-\\ Ended\end{tabular}} & \multicolumn{3}{c}{\textbf{RSVQA-HR Test Set 1}}    \\ \cmidrule(r){3-5}
&  & Presence & Comparison & Average Accuracy  \\ \hline
\multicolumn{5}{l}{\textit{\textbf{Specialist Models}: representative or SoTA methods with results reported in the literature}} \\
EasyToHard \cite{9771224} & \ding{55}  &  91.39 &  89.75 &  90.57  \\
SHRNet  \cite{10018408} & \ding{55}  & 92.45& 91.68 & 92.07\\ \hline
\multicolumn{5}{l}{\textit{\textbf{Generalist Models}: results of our own experimental runs (except RSGPT)}} \\
MiniGPT-4 \cite{zhu2023minigpt}&\ding{51}&52.91 &54.76 & 53.84\\
Shikra \cite{chen2023shikra} &\ding{51}&58.85 &57.40 &58.13 \\
MiniGPT-v2 \cite{chen2023minigptv2}&\ding{51} &64.80 &59.17 & 61.98 \\
RSGPT \cite{hu2023rsgpt} &\ding{55} & \textbf{91.86}  & \textbf{92.15}  & \textbf{92.00} \\
\rowcolor{violet!10}
$\text{SkyEyeGPT}_{single}$ &\ding{55}& \underline{87.59} &\underline{88.63} & \underline{88.11} \\ 
\rowcolor{violet!10}
$\text{SkyEyeGPT}_{one-stage}$ &\ding{51}& 83.19&83.92& 83.56 \\ 
\rowcolor{violet!10}
SkyEyeGPT &\ding{51}&  84.95&85.63 & 85.29 \\ \hline
\end{tabular}
}
\caption{\textbf{Results on RSVQA-HR test set 1.} Comparison of VQA results by Generalist and Specialist baselines.}
\label{RSVQAHR}
\end{table*}

\subsection*{C Additional Results}
\textbf{RS VQA:} We present the remote sensing VQA results for the RSVQA-HR test set 1 dataset, as shown in Table \ref{RSVQAHR}. The overall result trend is similar to Table 6 of the main paper.
% \ref{RSVQA}.
The model $\text{SkyEyeGPT}_{single}$ trained separately on the remote sensing VQA task outperforms the open-ended model, likely due to modality differences between satellite imagery in the RSVQA dataset and aerial imagery in other tasks. Sharing the same visual encoder across different modalities can lead to a performance loss for SkyEyeGPT. How to address this issue will be the focus of our next research work.

\textbf{RS Captioning:} Experimental results for evaluating remote sensing image captioning on Sydney-caption and RSICD datasets are provided in Tables \ref{Sydneycaption} and \ref{RSICDcaption}. We achieve the best results on both the UCM-caption (Table 3 of the main paper), Sydney-caption, and RSICD datasets. 
Even though all other metrics are optimal, CIDEr consistently performs worse than RSGPT. Through analysis, we find that our model tends to generate captions with different lengths or captions with richer semantics compared to the ground truth. Existing caption evaluation metrics struggle to provide accurate assessments, especially CIDEr.

\textbf{A novel evaluation method based on ChatGPT for remote sensing image captioning.}
To solve this issue, we utilize ChatGPT to determine whether the generated captions cover all visual objects and relations in the ground truth. We use ChatGPT to determine whether the generated caption is capable of being an alternative caption of the ground truth. For the Sydney-caption dataset, we randomly choose one ground-truth caption and treat it as the reference caption. We apply the following two prompts to perform the evaluation. The evaluation results of ChatGPT are shown in Table \ref{Sydneycaption_chatgpt}.

\textbf{Prompt 1:} \textit{There is one remote sensing image caption1 ‘{ground-truth caption}’, and there is another remote sensing image caption2 ‘{generated caption}’. Does remote sensing image caption2 cover all the objects and visual relations shown in remote sensing image caption1? Only answer yes or no without any explanation.}

\textbf{Prompt 2:} \textit{There is one remote sensing image caption1 ‘{ground-truth caption}’, and there is another remote sensing image caption2 ‘{generated caption}’. Based on remote sensing image caption1 and your understanding, do you think remote sensing image caption2 can be used as another caption? Only answer yes or no without any explanation.}

\begin{table}[h]
\centering
\scalebox{0.94}{
\begin{tabular}{lcc}
\hline
Method     & Accuracy 1 & Accuracy 2\\ \hline
MiniGPT-4  & 31.03\% &  46.55\% \\
Shirka     & 25.86\% &  43.10\%  \\
MiniGPT-v2 & 18.97\% &  32.76\%  \\
\rowcolor{violet!10}
\textbf{SkyEyeGPT}  & \textbf{51.72\%} & \textbf{56.90\%} \\ \hline
\end{tabular}
}
\caption{Sydney-caption evaluation using ChatGPT.}
\label{Sydneycaption_chatgpt}
\end{table}

The experimental results indicate that the captions generated by SkyEyeGPT are closer to the real visual objects and relations. The success accuracy of SkyEyeGPT is 51.72\% and 56.90\%, which is significantly higher than other methods. According to the traditional evaluation metrics in Table \ref{Sydneycaption}, MiniGPT-4 is the worst and MiniGPT-v2 is the best. From the results of GhatGPT, MiniGPT-4 is the best and MiniGPT-v2 is the worst. Therefore, it is necessary to rethink the evaluation method of the captioning task. Evaluating the captioning task based on ChatGPT may be a novel and more reasonable approach.

\begin{table*}[t]
\centering
\scalebox{0.9}{
\begin{tabular}{lcccccccc}
\hline
 \textbf{Method} & Open-Ended   & BLEU-1 & BLEU-2 & BLEU-3 & BLEU-4 & METEOR & ROUGE\_L & CIDEr  \\ \hline
\multicolumn{9}{l}{\textit{\textbf{Specialist Models}: representative or SoTA methods with results reported in the literature}} \\
SAA \cite{8931249}&\ding{55}&  68.82&60.73 &52.94 &45.39 &30.49& 58.20 &170.52   \\
Post-processing \cite{10138597} &\ding{55} & 78.37 & 69.85 & 63.22 & 57.17 &39.49   & 71.06 & {255.53} \\  \hline
\multicolumn{9}{l}{\textit{\textbf{Generalist Models}: results of our own experimental runs (except RSGPT)}} \\
MiniGPT-4 \cite{zhu2023minigpt}&\ding{51} & 29.53  & 25.85  & 20.27  & 16.38  & 32.02  & 42.73 & 0.07  \\
Shikra \cite{chen2023shikra}&\ding{51}&77.52&53.19&36.98& 27.82 &29.42 &53.27&26.79\\
MiniGPT-v2 \cite{chen2023minigptv2}&\ding{51} & 77.35  & 55.81  & 40.58  & 32.31  & 29.92  & 52.13    & 33.78  \\
RSGPT \cite{hu2023rsgpt} &\ding{55} &82.26 & 75.28  & 68.57 & 62.23 & 41.37 & 74.77 & \textbf{273.08} \\  
% \rowcolor{blue!7} 
\rowcolor{violet!10}
$\text{SkyEyeGPT}_{single}$ &\ding{55} &90.89 & 83.76 & 78.32 &74.15 &44.73 & 75.62 &\underline{191.46} \\
\rowcolor{violet!10}
$\text{SkyEyeGPT}_{one-stage}$ &\ding{51} &\underline{91.62} & \underline{85.33} & \underline{80.48} &\underline{76.75} &\underline{45.61} & \textbf{77.81} &173.98 \\
\rowcolor{violet!10}
SkyEyeGPT &\ding{51}  &\textbf{91.85} & \textbf{85.64} & \textbf{80.88} &\textbf{77.40} &\textbf{46.62} & \underline{77.74} &{181.06} \\ \hline
\end{tabular}
}
\caption{Comparisons with Generalist and Specialist models on \textbf{Sydney-captions dataset} for RS image captioning task.}
\label{Sydneycaption}
\end{table*}

\begin{table*}[]
\centering
\scalebox{0.9}{
\begin{tabular}{lcccccccc}
\hline
 \textbf{Method}  & Open-Ended & BLEU-1 & BLEU-2 & BLEU-3 & BLEU-4 & METEOR & ROUGE\_L & CIDEr  \\ \hline
\multicolumn{9}{l}{\textit{\textbf{Specialist Models}: representative or SoTA methods with results reported in the literature}} \\
SAA \cite{8931249}&\ding{55} &  59.35 &  45.11  &35.29   &28.08 &26.11& 49.57 &{132.35}   \\ 
Post-processing \cite{10138597}  &\ding{55}& 62.90  & 45.99 &35.68 & 28.68 &25.30   & 47.34 & 75.56 \\  \hline
\multicolumn{9}{l}{\textit{\textbf{Generalist Models}: results of our own experimental runs (except RSGPT)}} \\
MiniGPT-4 \cite{zhu2023minigpt}& \ding{51}& 33.98& 31.80& 25.83& 20.60&33.21& 40.72& 0.09  \\
Shikra \cite{chen2023shikra}&\ding{51}&82.61&62.51 &45.18&34.58&30.26&53.55&19.89\\
MiniGPT-v2 \cite{chen2023minigptv2}& \ding{51}& 83.10  & 64.55  & 47.83 & 37.28  & 30.21  & 54.00    & 52.41  \\
RSGPT \cite{hu2023rsgpt} &\ding{55}&70.32 & 54.23  & 44.02 & 36.83 & 30.10 & 53.34 & \textbf{102.94}\\ 
% \rowcolor{blue!7} 
\rowcolor{violet!10}
$\text{SkyEyeGPT}_{single}$ &\ding{55}&\textbf{87.33} & \textbf{77.70} & \textbf{68.90} &\textbf{61.99} &\textbf{36.23} & \textbf{63.54} &\underline{89.37} \\
\rowcolor{violet!10}
$\text{SkyEyeGPT}_{one-stage}$ &\ding{51} & 83.14 & 74.29 & 66.04 & 59.60&34.49&\underline{62.86}  &82.94 \\
\rowcolor{violet!10}
SkyEyeGPT & \ding{51} &\underline{86.71} & \underline{76.66}  & \underline{67.31} & \underline{59.99} & \underline{35.35} & {62.63} & 83.65\\ \hline
\end{tabular}
}
\caption{Comparisons with Generalist and Specialist models on \textbf{RSICD dataset} for RS image captioning task.}
\label{RSICDcaption}
\end{table*}

\subsection*{D Additional Ablation Studies}
\textbf{Vision-Language Alignment Layer.}
To demonstrate that our vision-language alignment layer is sufficient to align remote sensing visual and textual features, we design three variants: (a) removing the linear layer directly, (b) using multiple linear layers (two or three) instead of one linear layer, and (c) using a Q-Former instead of one linear layer. All variants are trained the same way. The result is as shown in Table \ref{vis_pro}. None of the variants performed very well, with the (c) SkyEyeGPT + Q-Former is closest to SkyEyeGPT.
This shows that under the training of our SkyEye-958k instruction, a single linear layer is sufficient to align remote sensing visual features with LLM.

\begin{table*}[b]
\centering
\scalebox{0.95}{
\begin{tabular}{lcccc}
\hline
\textbf{Model} & \textbf{UCM-caption} & \textbf{CapERA} & \textbf{DIOR-RSVG} & \textbf{RSVQA-LR} \\ \hline
(a) SkyEyeGPT w/o Linear Layer  & 38.35 & 29.91 & 47.12 & 42.67  \\
(b) SkyEyeGPT + 2 Linear Layers  & 167.92 & 55.35 & 66.21  & 51.88 \\
(b) SkyEyeGPT + 3 Linear Layers & 90.22 & 40.77 &  52.78& 43.11 \\
(c) SkyEyeGPT + Q-Former  & 181.39&  71.68 & 63.59&  58.66  \\ 
\rowcolor{violet!10}
\textbf{SkyEyeGPT} & \textbf{236.75} & \textbf{91.90} & \textbf{88.59} & \textbf{84.19} \\
 SkyEyeGPT w/o Task Identifier &208.46 & 84.15 & 79.04  & 80.28 \\ \hline
\end{tabular}
}
\caption{Ablation on designs of the vision-language alignment layer and task identifier.}
\label{vis_pro}
\end{table*}

\textbf{Task Identifier.} 
In order to study the impact of task identifiers on the results, we conducted an ablation experiment, and the results are shown in Table \ref{vis_pro}. SkyEyeGPT w/o Task Identifier indicates that the task identifier in all instructions is removed for training. The results on four tasks validate the clear advantages of adding task identifiers. Task identifier is conducive to improving the efficiency of multi-task learning and improving SkyEyeGPT's performance on each task.

\textbf{Low-Rank Adaptation (LoRA).} 
In order to study the effect of the rank of LoRA on the results, we set different ranks for experiments. The results are shown in Table \ref{lora_rank}. With the increase of rank, the model performance first increases and then decreases. The model performance is best when the rank is 64.

\begin{table}[h]
\centering
\scalebox{0.84}{
\begin{tabular}{ccccc}
\hline
\textbf{rank} & \textbf{UCM-caption} & \textbf{CapERA} & \textbf{DIOR-RSVG} & \textbf{RSVQA-LR} \\ \hline
16  & 71.66 & 35.17 & 43.22 & 45.71  \\
32  & 112.50 & 44.28 & 56.48  & 60.42 \\
\rowcolor{violet!10}
\textbf{64} & \textbf{236.75} & \textbf{91.90} & \textbf{88.59} & \textbf{84.19} \\
96 & 123.55 & 56.13 & 61.27  & 64.39 \\
128 & 64.88& 33.64 & 41.91  &42.11  \\
\hline
\end{tabular}
}
\caption{Ablation on the rank of LoRA.}
\label{lora_rank}
\end{table}

\subsection*{E More Qualitative Analysis}
\begin{figure*}
  \centering
  \includegraphics[width=0.95\linewidth]{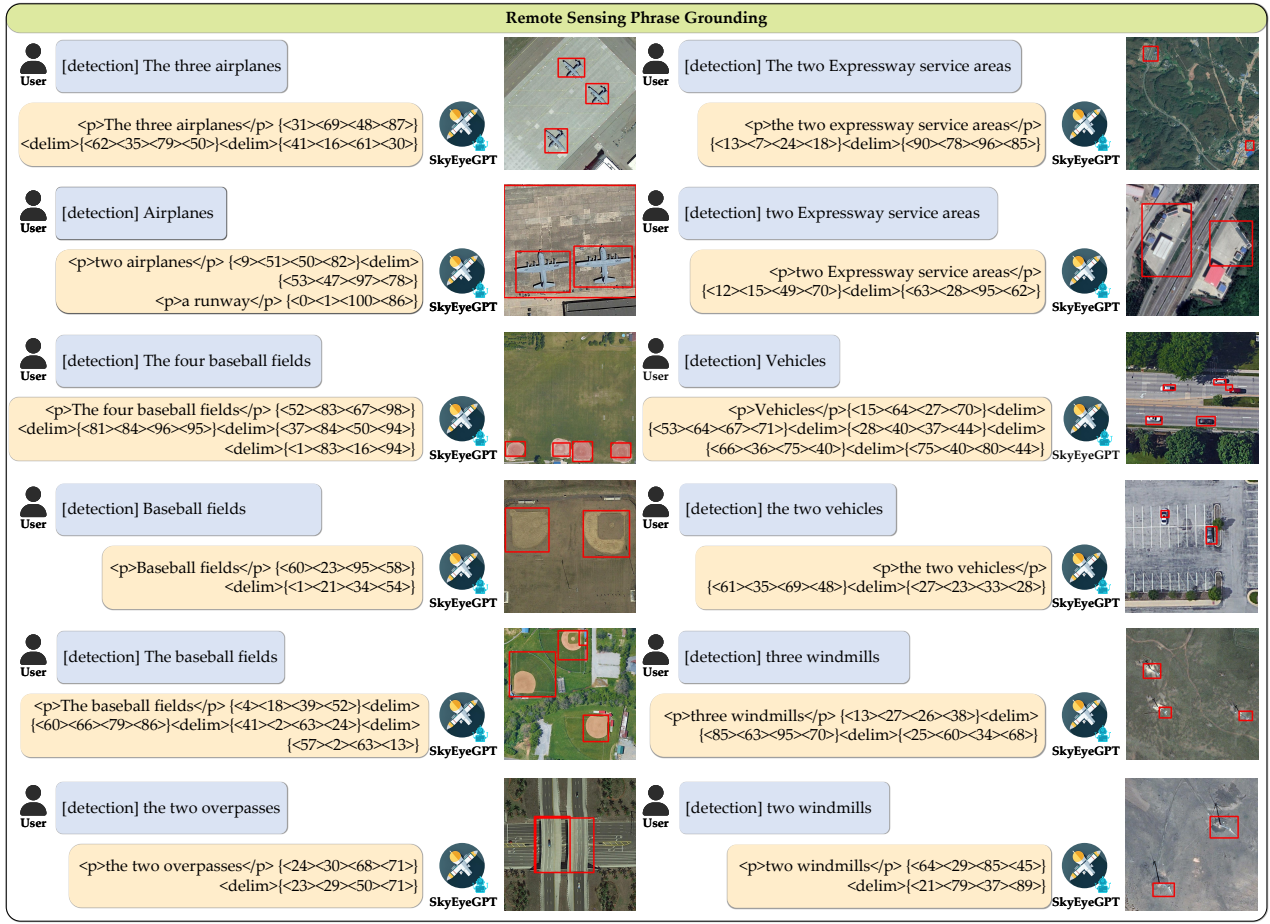}
  \caption{
  Visualization results of \textbf{remote sensing phrase grounding} by SkyEyeGPT.
  }
  \label{fig:more_phrasegrounding}
\end{figure*}

\begin{figure*}
  \centering
  \includegraphics[width=0.93\linewidth]{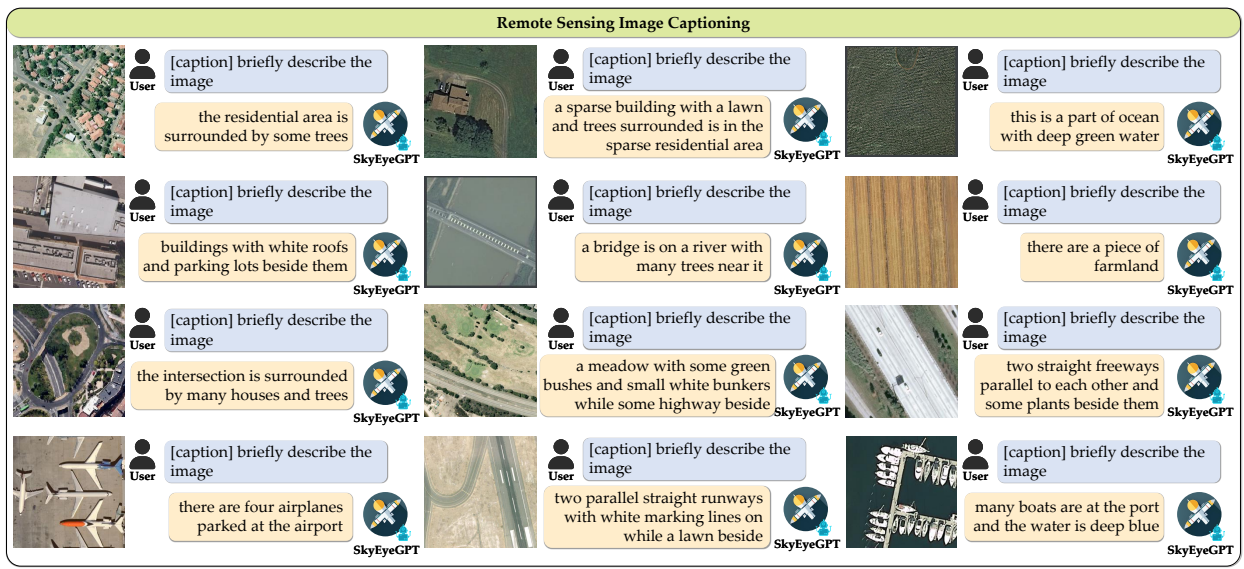}
  \caption{
  Visualization results of \textbf{remote sensing image captioning} by SkyEyeGPT.
  }
  \label{fig:more_captioning}
\end{figure*}

\begin{figure*}
  \centering
  \includegraphics[width=0.9\linewidth]{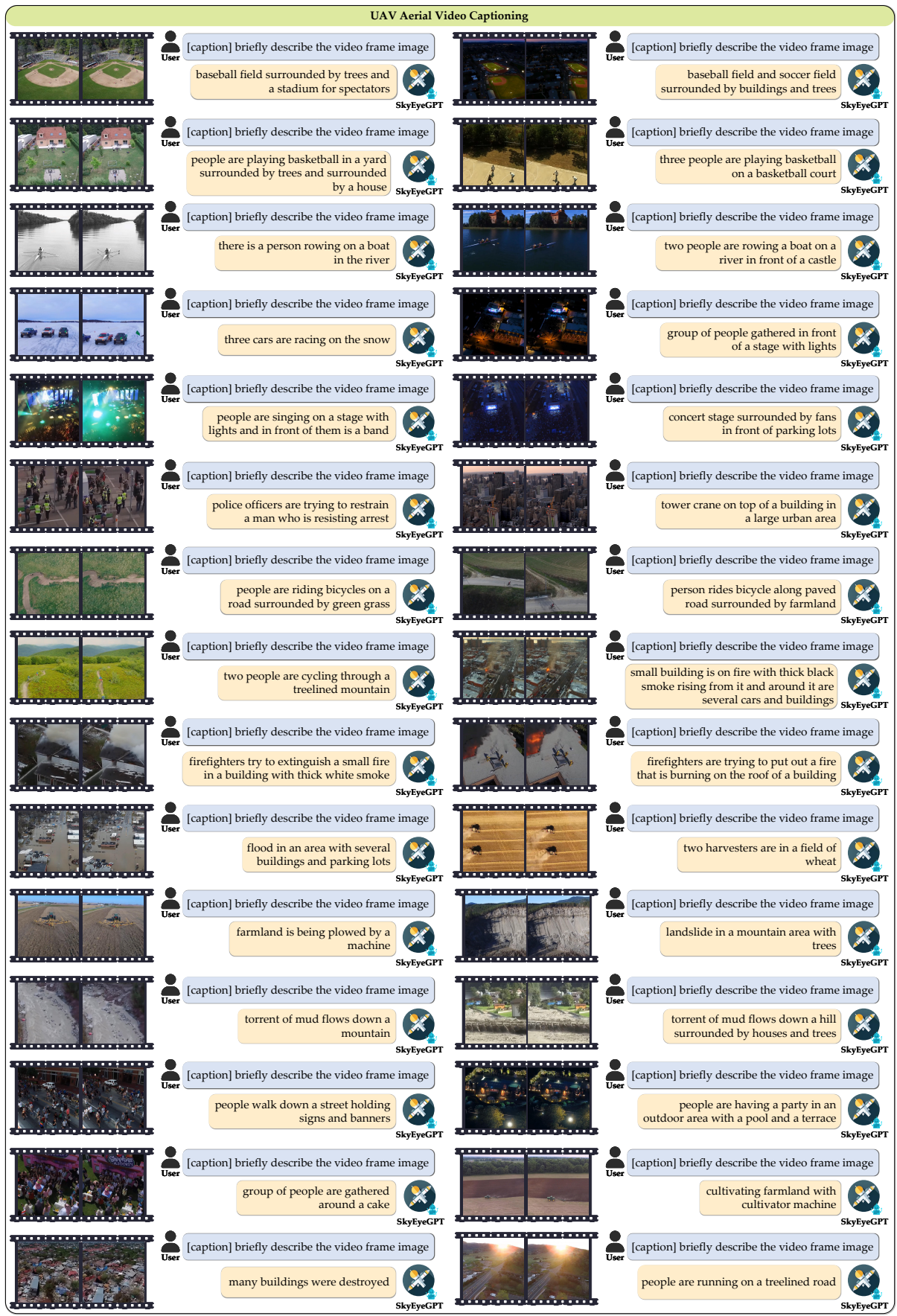}
  \caption{
  Visualization results of \textbf{UAV aerial video captioning} by SkyEyeGPT.
  }
  \label{fig:more_UAVcaptioning}
\end{figure*}

\begin{figure*}
  \centering
  \includegraphics[width=0.95\linewidth]{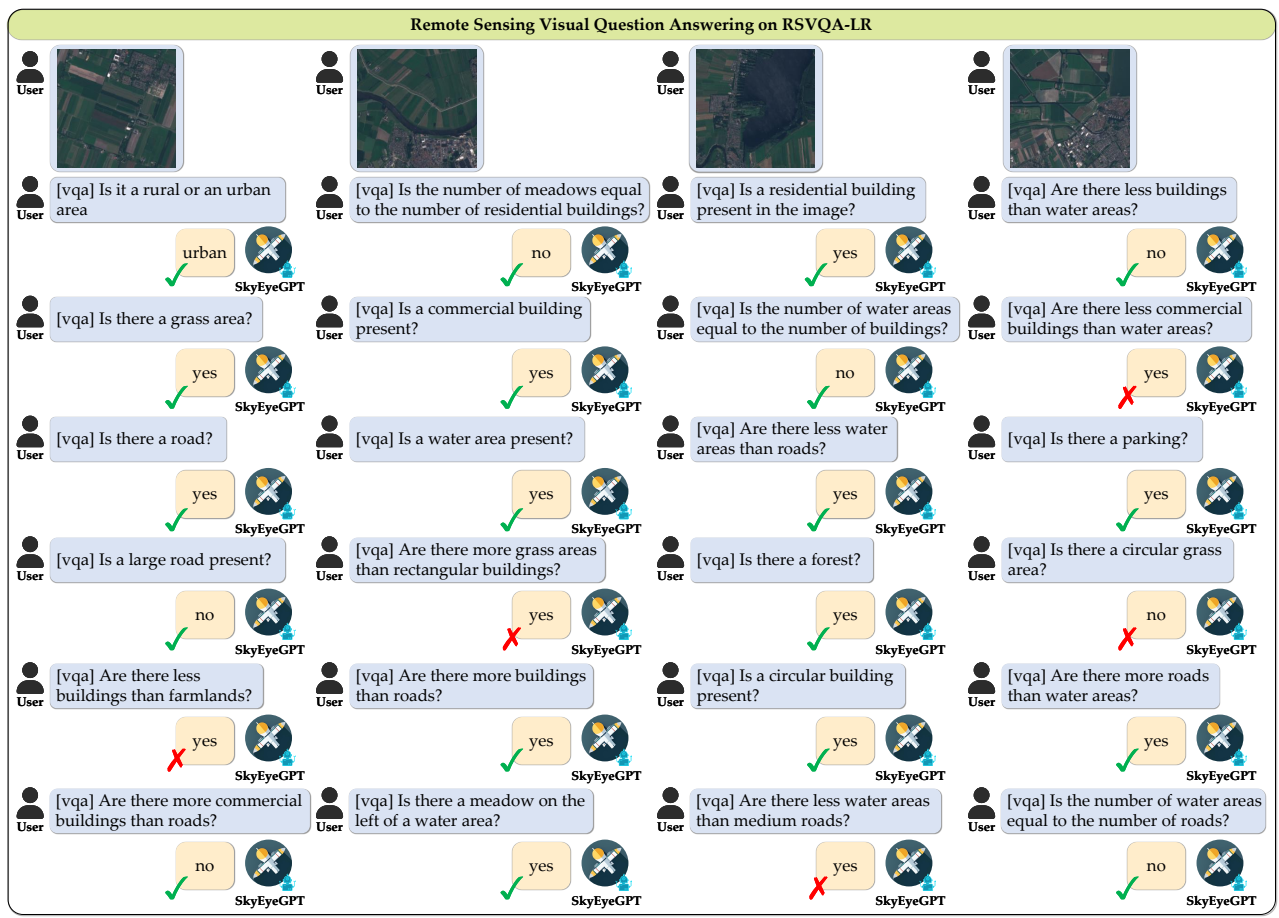}
  \caption{
  Visualization results of \textbf{remote sensing visual question answering} by SkyEyeGPT.
  }
  \label{fig:more_vqa}
\end{figure*}

\begin{figure*}
  \centering
  \includegraphics[width=0.95\linewidth]{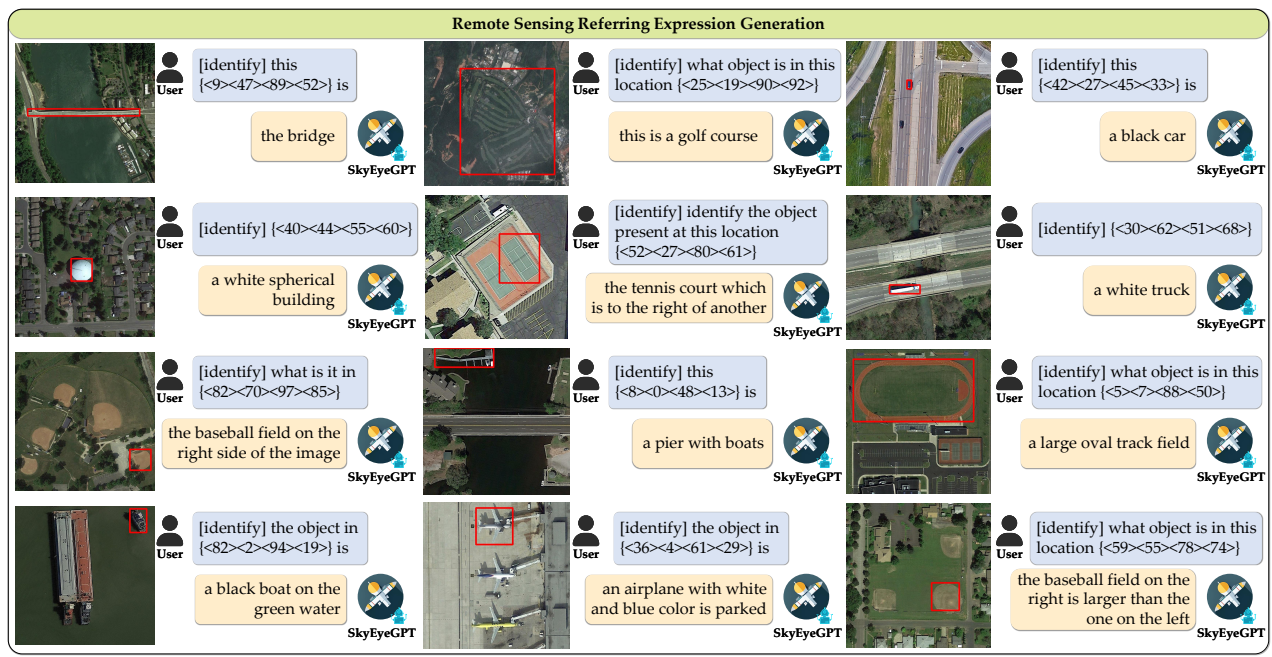}
  \caption{
  Visualization results of \textbf{remote sensing referring expression generation} by SkyEyeGPT.
  }
  \label{fig:more_generate}
\end{figure*}

\begin{figure*}
  \centering
  \includegraphics[width=0.95\linewidth]{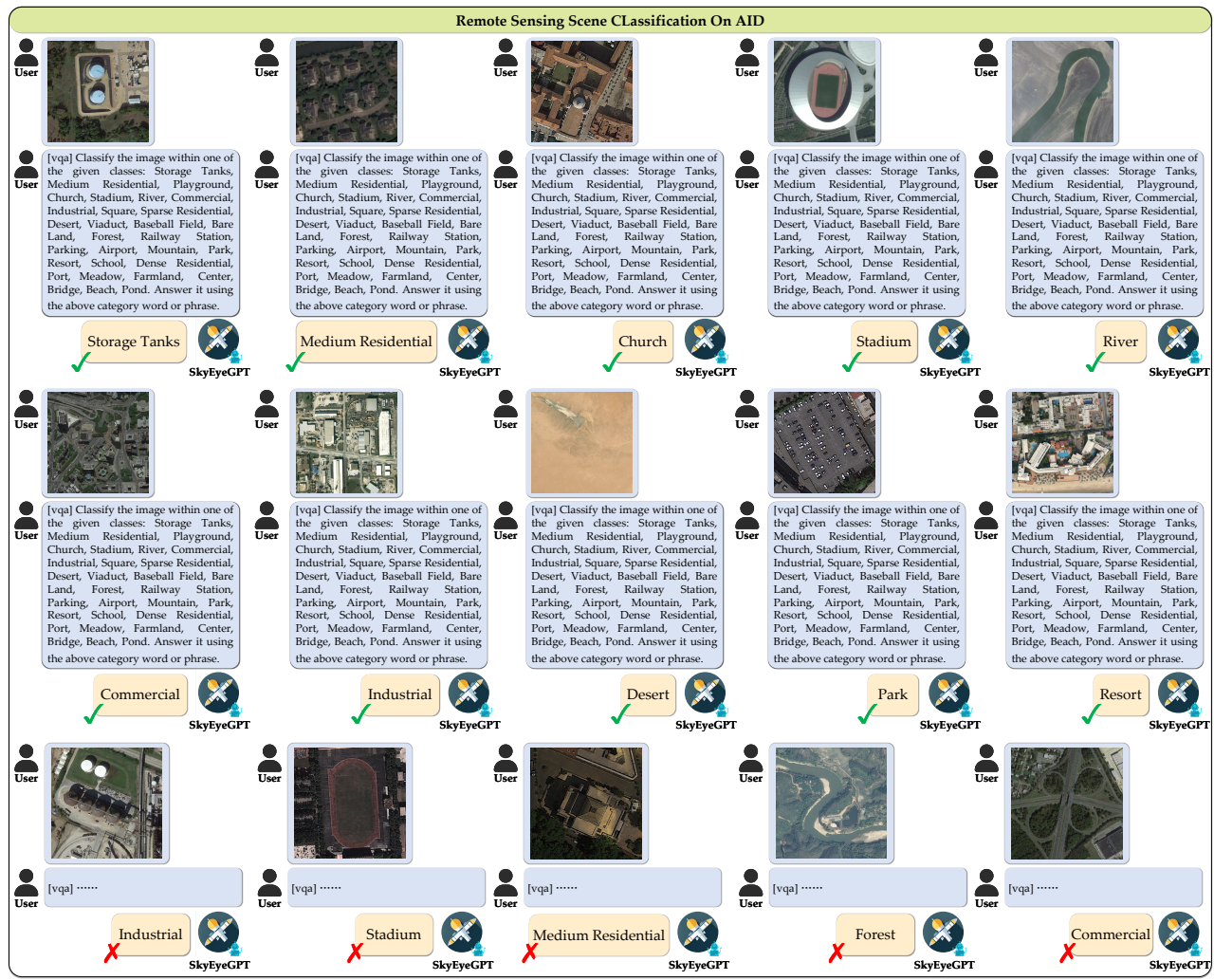}
  \caption{
  Visualization results of \textbf{remote sensing scene classification} by SkyEyeGPT.
  }
  \label{fig:more_vqaclass}
\end{figure*}

We provide additional qualitative results from SkyEyeGPT in Figures \ref{fig:more_grounding}, \ref{fig:more_phrasegrounding}, \ref{fig:more_captioning}, \ref{fig:more_UAVcaptioning}, \ref{fig:more_vqa}, \ref{fig:more_generate}, and \ref{fig:more_vqaclass}.

% \ref{fig:more_conversa}.

% \textbf{Multi-round multi-task conversational ability of SkyEyeGPT.} 
 
\begin{figure*}
  \centering
  \includegraphics[width=0.95\linewidth]{10-more_grounding_results.pdf}
  \caption{
  Visualization results of \textbf{remote sensing visual grounding} by SkyEyeGPT.
  }
  \label{fig:more_grounding}
\end{figure*}

\bibliographystyle{named}
\bibliography{ijcai24}

\begin{thebibliography}{}

\bibitem[\protect\citeauthoryear{Bashmal \bgroup \em et al.\egroup }{2023}]{rs15082139}
Laila Bashmal, Yakoub Bazi, Mohamad~Mahmoud Al~Rahhal, Mansour Zuair, and Farid Melgani.
\newblock Capera: Captioning events in aerial videos.
\newblock {\em Remote Sensing}, 15(8), 2023.

\bibitem[\protect\citeauthoryear{Chen \bgroup \em et al.\egroup }{2023a}]{chen2023minigptv2}
Jun Chen, Deyao Zhu, Xiaoqian Shen, Xiang Li, Zechun Liu, Pengchuan Zhang, Raghuraman Krishnamoorthi, Vikas Chandra, Yunyang Xiong, and Mohamed Elhoseiny.
\newblock Minigpt-v2: large language model as a unified interface for vision-language multi-task learning.
\newblock {\em arXiv preprint arXiv:2310.09478}, 2023.

\bibitem[\protect\citeauthoryear{Chen \bgroup \em et al.\egroup }{2023b}]{chen2023shikra}
Keqin Chen, Zhao Zhang, Weili Zeng, Richong Zhang, Feng Zhu, and Rui Zhao.
\newblock Shikra: Unleashing multimodal llm's referential dialogue magic.
\newblock {\em arXiv preprint arXiv:2306.15195}, 2023.

\bibitem[\protect\citeauthoryear{Cheng \bgroup \em et al.\egroup }{2022}]{9866055}
Qimin Cheng, Haiyan Huang, Yuan Xu, Yuzhuo Zhou, Huanying Li, and Zhongyuan Wang.
\newblock Nwpu-captions dataset and mlca-net for remote sensing image captioning.
\newblock {\em IEEE Transactions on Geoscience and Remote Sensing}, 60:1--19, 2022.

\bibitem[\protect\citeauthoryear{Chiang \bgroup \em et al.\egroup }{2023}]{chiang2023vicuna}
Wei-Lin Chiang, Zhuohan Li, Zi~Lin, Ying Sheng, Zhanghao Wu, Hao Zhang, Lianmin Zheng, Siyuan Zhuang, Yonghao Zhuang, Joseph~E Gonzalez, et~al.
\newblock Vicuna: An open-source chatbot impressing gpt-4 with 90\%* chatgpt quality.
\newblock {\em See https://vicuna. lmsys. org (accessed 14 April 2023)}, 2023.

\bibitem[\protect\citeauthoryear{Dai \bgroup \em et al.\egroup }{2023}]{instructblip}
Wenliang Dai, Junnan Li, Dongxu Li, Anthony Meng~Huat Tiong, Junqi Zhao, Weisheng Wang, Boyang Li, Pascale Fung, and Steven Hoi.
\newblock Instructblip: Towards general-purpose vision-language models with instruction tuning.
\newblock {\em arXiv preprint arXiv:2305.06500}, 2023.

\bibitem[\protect\citeauthoryear{Fang \bgroup \em et al.\egroup }{2023}]{Fang_2023_CVPR}
Yuxin Fang, Wen Wang, Binhui Xie, Quan Sun, Ledell Wu, Xinggang Wang, Tiejun Huang, Xinlong Wang, and Yue Cao.
\newblock Eva: Exploring the limits of masked visual representation learning at scale.
\newblock In {\em Proceedings of the IEEE/CVF Conference on Computer Vision and Pattern Recognition (CVPR)}, pages 19358--19369, June 2023.

\bibitem[\protect\citeauthoryear{Hoxha \bgroup \em et al.\egroup }{2023}]{10138597}
Genc Hoxha, Giacomo Scuccato, and Farid Melgani.
\newblock Improving image captioning systems with postprocessing strategies.
\newblock {\em IEEE Transactions on Geoscience and Remote Sensing}, 61:1--13, 2023.

\bibitem[\protect\citeauthoryear{Hu \bgroup \em et al.\egroup }{2023}]{hu2023rsgpt}
Yuan Hu, Jianlong Yuan, Congcong Wen, Xiaonan Lu, and Xiang Li.
\newblock Rsgpt: A remote sensing vision language model and benchmark.
\newblock {\em arXiv preprint arXiv:2307.15266}, 2023.

\bibitem[\protect\citeauthoryear{Huang \bgroup \em et al.\egroup }{2021}]{Huang_2021_CVPR}
Binbin Huang, Dongze Lian, Weixin Luo, and Shenghua Gao.
\newblock Look before you leap: Learning landmark features for one-stage visual grounding.
\newblock In {\em Proceedings of the IEEE/CVF Conference on Computer Vision and Pattern Recognition (CVPR)}, pages 16888--16897, June 2021.

\bibitem[\protect\citeauthoryear{Li \bgroup \em et al.\egroup }{2020}]{li2020object}
Ke~Li, Gang Wan, Gong Cheng, Liqiu Meng, and Junwei Han.
\newblock Object detection in optical remote sensing images: A survey and a new benchmark.
\newblock {\em ISPRS journal of photogrammetry and remote sensing}, 159:296--307, 2020.

\bibitem[\protect\citeauthoryear{Li \bgroup \em et al.\egroup }{2023}]{li2023blip}
Junnan Li, Dongxu Li, Silvio Savarese, and Steven Hoi.
\newblock Blip-2: Bootstrapping language-image pre-training with frozen image encoders and large language models.
\newblock {\em arXiv preprint arXiv:2301.12597}, 2023.

\bibitem[\protect\citeauthoryear{Liu \bgroup \em et al.\egroup }{2023}]{liu2023visual}
Haotian Liu, Chunyuan Li, Qingyang Wu, and Yong~Jae Lee.
\newblock Visual instruction tuning.
\newblock In {\em NeurIPS}, 2023.

\bibitem[\protect\citeauthoryear{Lobry \bgroup \em et al.\egroup }{2020}]{9088993}
Sylvain Lobry, Diego Marcos, Jesse Murray, and Devis Tuia.
\newblock Rsvqa: Visual question answering for remote sensing data.
\newblock {\em IEEE Transactions on Geoscience and Remote Sensing}, 58(12):8555--8566, 2020.

\bibitem[\protect\citeauthoryear{Lu \bgroup \em et al.\egroup }{2018}]{8240966}
Xiaoqiang Lu, Binqiang Wang, Xiangtao Zheng, and Xuelong Li.
\newblock Exploring models and data for remote sensing image caption generation.
\newblock {\em IEEE Transactions on Geoscience and Remote Sensing}, 56(4):2183--2195, 2018.

\bibitem[\protect\citeauthoryear{Lu \bgroup \em et al.\egroup }{2020}]{8931249}
Xiaoqiang Lu, Binqiang Wang, and Xiangtao Zheng.
\newblock Sound active attention framework for remote sensing image captioning.
\newblock {\em IEEE Transactions on Geoscience and Remote Sensing}, 58(3):1985--2000, 2020.

\bibitem[\protect\citeauthoryear{Mou \bgroup \em et al.\egroup }{2020}]{9295448}
Lichao Mou, Yuansheng Hua, Pu~Jin, and Xiao~Xiang Zhu.
\newblock Era: A data set and deep learning benchmark for event recognition in aerial videos [software and data sets].
\newblock {\em IEEE Geoscience and Remote Sensing Magazine}, 8(4):125--133, 2020.

\bibitem[\protect\citeauthoryear{OpenAI}{2022}]{Chatgpt}
OpenAI.
\newblock Chatgpt.
\newblock \url{https://openai.com/blog/chatgpt}, 2022.

\bibitem[\protect\citeauthoryear{OpenAI}{2023}]{Gpt4}
OpenAI.
\newblock Gpt-4 technical report.
\newblock {\em arXiv}, 2023.

\bibitem[\protect\citeauthoryear{Qu \bgroup \em et al.\egroup }{2016}]{7546397}
Bo~Qu, Xuelong Li, Dacheng Tao, and Xiaoqiang Lu.
\newblock Deep semantic understanding of high resolution remote sensing image.
\newblock In {\em 2016 International Conference on Computer, Information and Telecommunication Systems (CITS)}, pages 1--5, 2016.

\bibitem[\protect\citeauthoryear{Radford \bgroup \em et al.\egroup }{2021}]{radford2021learning}
Alec Radford, Jong~Wook Kim, Chris Hallacy, Aditya Ramesh, Gabriel Goh, Sandhini Agarwal, Girish Sastry, Amanda Askell, Pamela Mishkin, Jack Clark, et~al.
\newblock Learning transferable visual models from natural language supervision.
\newblock In {\em International conference on machine learning}, pages 8748--8763, 2021.

\bibitem[\protect\citeauthoryear{Sun \bgroup \em et al.\egroup }{2022}]{3548316}
Yuxi Sun, Shanshan Feng, Xutao Li, Yunming Ye, Jian Kang, and Xu~Huang.
\newblock Visual grounding in remote sensing images.
\newblock In {\em Proceedings of the 30th ACM International Conference on Multimedia}, pages 404--412, 2022.

\bibitem[\protect\citeauthoryear{Touvron \bgroup \em et al.\egroup }{2023a}]{touvron2023llama}
Hugo Touvron, Thibaut Lavril, Gautier Izacard, Xavier Martinet, Marie-Anne Lachaux, Timothée Lacroix, Baptiste Rozière, Naman Goyal, Eric Hambro, Faisal Azhar, Aurelien Rodriguez, Armand Joulin, Edouard Grave, and Guillaume Lample.
\newblock Llama: Open and efficient foundation language models.
\newblock {\em arXiv preprint arXiv:2302.13971}, 2023.

\bibitem[\protect\citeauthoryear{Touvron \bgroup \em et al.\egroup }{2023b}]{touvron2023llama2}
Hugo Touvron, Louis Martin, Kevin Stone, Peter Albert, Amjad Almahairi, Yasmine Babaei, Nikolay Bashlykov, Soumya Batra, Prajjwal Bhargava, Shruti Bhosale, et~al.
\newblock Llama 2: Open foundation and fine-tuned chat models.
\newblock {\em arXiv preprint arXiv:2307.09288}, 2023.

\bibitem[\protect\citeauthoryear{Wang \bgroup \em et al.\egroup }{2023}]{wang2023visionllm}
Wenhai Wang, Zhe Chen, Xiaokang Chen, Jiannan Wu, Xizhou Zhu, Gang Zeng, Ping Luo, Tong Lu, Jie Zhou, Yu~Qiao, et~al.
\newblock Visionllm: Large language model is also an open-ended decoder for vision-centric tasks.
\newblock {\em arXiv preprint arXiv:2305.11175}, 2023.

\bibitem[\protect\citeauthoryear{Xia \bgroup \em et al.\egroup }{2018}]{xia2018dota}
Gui-Song Xia, Xiang Bai, Jian Ding, Zhen Zhu, Serge Belongie, Jiebo Luo, Mihai Datcu, Marcello Pelillo, and Liangpei Zhang.
\newblock Dota: A large-scale dataset for object detection in aerial images.
\newblock In {\em Proceedings of the IEEE conference on computer vision and pattern recognition}, pages 3974--3983, 2018.

\bibitem[\protect\citeauthoryear{Xiong \bgroup \em et al.\egroup }{2022}]{xiong2022earthnets}
Zhitong Xiong, Fahong Zhang, Yi~Wang, Yilei Shi, and Xiao~Xiang Zhu.
\newblock Earthnets: Empowering {AI} in earth observation.
\newblock {\em arXiv preprint arXiv:2210.04936}, 2022.

\bibitem[\protect\citeauthoryear{Xu \bgroup \em et al.\egroup }{2023}]{xu2023u}
Jinjin Xu, Liwu Xu, Yuzhe Yang, Xiang Li, Yanchun Xie, Yi-Jie Huang, and Yaqian Li.
\newblock u-llava: Unifying multi-modal tasks via large language model.
\newblock {\em arXiv preprint arXiv:2311.05348}, 2023.

\bibitem[\protect\citeauthoryear{Yang \bgroup \em et al.\egroup }{2019}]{yang2019fast}
Zhengyuan Yang, Boqing Gong, Liwei Wang, Wenbing Huang, Dong Yu, and Jiebo Luo.
\newblock A fast and accurate one-stage approach to visual grounding.
\newblock In {\em Proceedings of the IEEE/CVF International Conference on Computer Vision}, pages 4683--4693, 2019.

\bibitem[\protect\citeauthoryear{Yang \bgroup \em et al.\egroup }{2020}]{yang2020improving}
Zhengyuan Yang, Tianlang Chen, Liwei Wang, and Jiebo Luo.
\newblock Improving one-stage visual grounding by recursive sub-query construction.
\newblock In {\em Computer Vision--ECCV 2020: 16th European Conference, Glasgow, UK, August 23--28, 2020, Proceedings, Part XIV 16}, pages 387--404, 2020.

\bibitem[\protect\citeauthoryear{Yuan \bgroup \em et al.\egroup }{2022a}]{9771224}
Zhenghang Yuan, Lichao Mou, Qi~Wang, and Xiao~Xiang Zhu.
\newblock From easy to hard: Learning language-guided curriculum for visual question answering on remote sensing data.
\newblock {\em IEEE Transactions on Geoscience and Remote Sensing}, 60:1--11, 2022.

\bibitem[\protect\citeauthoryear{Yuan \bgroup \em et al.\egroup }{2022b}]{9437331}
Zhiqiang Yuan, Wenkai Zhang, Kun Fu, Xuan Li, Chubo Deng, Hongqi Wang, and Xian Sun.
\newblock Exploring a fine-grained multiscale method for cross-modal remote sensing image retrieval.
\newblock {\em IEEE Transactions on Geoscience and Remote Sensing}, 60:1--19, 2022.

\bibitem[\protect\citeauthoryear{Yuan \bgroup \em et al.\egroup }{2023a}]{10231134}
Yuan Yuan, Yang Zhan, and Zhitong Xiong.
\newblock Parameter-efficient transfer learning for remote sensing image–text retrieval.
\newblock {\em IEEE Transactions on Geoscience and Remote Sensing}, 61:1--14, 2023.

\bibitem[\protect\citeauthoryear{Yuan \bgroup \em et al.\egroup }{2023b}]{10282946}
Zhenghang Yuan, Lichao Mou, and Xiao~Xiang Zhu.
\newblock Overcoming language bias in remote sensing visual question answering via adversarial training.
\newblock In {\em IGARSS 2023 - 2023 IEEE International Geoscience and Remote Sensing Symposium}, pages 2235--2238, 2023.

\bibitem[\protect\citeauthoryear{Zhan \bgroup \em et al.\egroup }{2023a}]{10056343}
Yang Zhan, Zhitong Xiong, and Yuan Yuan.
\newblock Rsvg: Exploring data and models for visual grounding on remote sensing data.
\newblock {\em IEEE Transactions on Geoscience and Remote Sensing}, 61:1--13, 2023.

\bibitem[\protect\citeauthoryear{Zhan \bgroup \em et al.\egroup }{2023b}]{zhan2023mono3dvg}
Yang Zhan, Yuan Yuan, and Zhitong Xiong.
\newblock Mono3dvg: 3d visual grounding in monocular images.
\newblock {\em arXiv preprint arXiv:2312.08022}, 2023.

\bibitem[\protect\citeauthoryear{Zhang \bgroup \em et al.\egroup }{2023}]{10018408}
Zixiao Zhang, Licheng Jiao, Lingling Li, Xu~Liu, Puhua Chen, Fang Liu, Yuxuan Li, and Zhicheng Guo.
\newblock A spatial hierarchical reasoning network for remote sensing visual question answering.
\newblock {\em IEEE Transactions on Geoscience and Remote Sensing}, 61:1--15, 2023.

\bibitem[\protect\citeauthoryear{Zheng \bgroup \em et al.\egroup }{2022}]{9444570}
Xiangtao Zheng, Binqiang Wang, Xingqian Du, and Xiaoqiang Lu.
\newblock Mutual attention inception network for remote sensing visual question answering.
\newblock {\em IEEE Transactions on Geoscience and Remote Sensing}, 60:1--14, 2022.

\bibitem[\protect\citeauthoryear{Zhu \bgroup \em et al.\egroup }{2023}]{zhu2023minigpt}
Deyao Zhu, Jun Chen, Xiaoqian Shen, Xiang Li, and Mohamed Elhoseiny.
\newblock Minigpt-4: Enhancing vision-language understanding with advanced large language models.
\newblock {\em arXiv preprint arXiv:2304.10592}, 2023.

\end{thebibliography}

\end{document}